\documentclass[10pt,twocolumn,letterpaper]{article}

\usepackage{iccv}
\usepackage{times}
\usepackage{epsfig}
\usepackage{graphicx}
\usepackage{amsmath}
\usepackage{amssymb}
\usepackage{bbm}
\usepackage{enumitem}
\usepackage{diagbox}
\usepackage{adjustbox, multirow}
\usepackage[ruled,vlined]{algorithm2e}
\usepackage{soul}
\usepackage{placeins}
\usepackage[dvipsnames]{xcolor}
\usepackage[pagebackref=true,breaklinks=true,letterpaper=true,colorlinks,bookmarks=false]{hyperref}

\iccvfinalcopy % *** Uncomment this line for the final submission

\newcommand{\yifan}[1]{{\color{BurntOrange} [[Yifan:``#1'']]}}
\newcommand{\tong}[1]{{}}
\newcommand{\tianjun}[1]{{}}
\newcommand{\david}[1]{{}}
\newcommand{\zz}[1]{{}}
\newcommand{\yuanjx}[1]{}
\newcommand{\note}[1]{{\color{red} #1}}
\newcommand{\updateblue}[1]{{\color{black}#1}}
\newcommand{\updateblueafter}[1]{{\color{black}#1}}

\newcommand{\updatered}[1]{{\color{black}#1}}
\newcommand{\updatepurple}[1]{{\color{black}#1}}

\newcommand{\NAME}{LANDER\xspace}
\newcommand{\HNAME}{Hi-LANDER\xspace}

\newcommand{\myfigref}[1]{Figure~\ref{#1}}
\newcommand{\mytabref}[1]{Table~\ref{#1}}
\newcommand{\mysecref}[1]{Section~\ref{#1}}

\makeatletter
\def\blfootnote{\xdef\@thefnmark{}\@footnotetext}
\makeatother

\def\iccvPaperID{8656} % *** Enter the ICCV Paper ID here

\ificcvfinal\fi

\begin{document}

%%%%%%%%% TITLE
\title{Learning Hierarchical Graph Neural Networks for Image Clustering} % arxiv
% \title{%Hierarchical Graph Neural Networks for Visual Clustering \\
%  \yifan{Visual} Meta-Clustering with \note{Hierarchical} Graph Neural Networks}
% \footnotemark[\value{footnote}]

% \author{Yifan Xing\footnotemark \and  Tong He\footnotemark[1] \and  Tianjun Xiao \and  Yongxin Wang \and  Yuanjun Xiong \and  Wei Xia \and  David Wipf Paul \and  Zheng Zhang \and Stefano Soatto \\
%   Amazon Web Services \\
%   {\tt\small \{yifax, htong, tianjux, yongxinw, yuanjx, wxia, daviwipf, zhaz, soattos\}@amazon.com}
% }

\author{Yifan Xing\footnotemark  \hfill  Tong He\footnotemark[1]  \hfill  Tianjun Xiao  \hfill  Yongxin Wang  \hfill  Yuanjun Xiong \\
  Wei Xia  \hfill  David Wipf  \hfill  Zheng Zhang  \hfill Stefano Soatto \\\\
  Amazon Web Services \\
  {\tt\small \{yifax, htong, tianjux, yongxinw, yuanjx, wxia, daviwipf, zhaz, soattos\}@amazon.com}
}

\maketitle

\blfootnote{*Indicates equal contribution.}

\maketitle
\ificcvfinal\thispagestyle{empty}\fi

% \note{I am afraid reviewers will see the title as too general, so adding ``hierarchical'' as a qualified}

% \yifan{What about adding visual? ``Hierarchical Visual Meta-Clustering with Graph Neural Networks'', to avoid reviewers 
%     potentially attack why not work on inputs other than images.}
    
% \yifan{Meta-clustering confusion with con-census clustering. Supervised clustering is the consistent notation with prior literature. Meta-clustering a terminology is already in use.}

% \note{I suggest changing the title to remove the first word. You can "learn clustering", and you can "train networks", but "learning networks" is not a sensible concept. Learning made sense in earlier titles about learning to cluster, that were dismissed, but makes no sense in this title.}

\begin{abstract}
%We propose a method to exploit a training set of images and corresponding labels to learn how to cluster a different test set of images for which no annotations are available. The number of clusters in the test set is unknown, and the set of labels is disjoint from that of the training set. 
We propose a hierarchical graph neural network (GNN) model that learns how to cluster a set of images into an unknown number of identities using a training set of images annotated with labels belonging to a disjoint set of identities. 
Our hierarchical GNN uses a novel approach to merge connected components predicted at each level of the hierarchy to form a new graph at the next level. Unlike fully unsupervised hierarchical clustering, the choice of grouping and complexity criteria stems naturally from supervision in the training set.  The resulting method, \HNAME, achieves an average of 54\% improvement in F-score and 8\% increase in Normalized Mutual Information (NMI) relative to current GNN-based clustering algorithms.  Additionally,  state-of-the-art 
% \note{always define an acronym when first used, and avoid as much as possible using acronyms in title and abstract} 
GNN-based methods rely on separate models to  predict linkage probabilities and node densities as intermediate steps of the clustering process. In contrast, our unified framework achieves a seven-fold decrease in computational cost. We release our training and inference code \href{https://github.com/dmlc/dgl/tree/master/examples/pytorch/hilander}{here}.
\end{abstract}

% \note{I am introducing the name in the abstract, since later you reference it several pages down. It is not clear what Hi-LANDER stands for, so I picked one more related to the title, but feel free to change it if Hi-LANDER has a meaning
% } \yifan{\NAME stands for Linkage Approximation aNd Density Estimation Refinement, the base GNN design. Then \HNAME refers to the hierarchical end-end method. I think it depends on how much emphasis we want to put in the novelty about \NAME module, the key differentiator to prior work is the joint inference and unified loss, concept of density and linkage are not new, maybe we want to emphasize joint and unified in the name. TO BE DISCUSSED.}

\section{Introduction}

Clustering is a pillar of unsupervised learning. It consists of grouping data points according to a manually specified criterion. Without any supervision, the problem is self-referential, with the outcome being defined by the choice of grouping criterion. Different criteria yield different solutions, with no independent validation mechanism. Even within a given criterion, clustering typically yields multiple solutions depending on a complexity measure, and a separate model selection criterion is introduced to arrive at a unique solution. A large branch of unsupervised clustering methods follow the hierarchical/agglomerative framework \cite{HAC, slonim1999agglomerative, quickshift}, \updateblueafter{which gives a tree of cluster partitions with varying granularity of the data, but they still require a model selection criterion for the final single grouping.} 
% \note{the previous sentence is problematic. "the issue" is missing a connection. What issue? Also, agglomerative methods do not allow more control, but rather produce the entire homotopy class of segmentations (or, if you prefer, the segmentation tree), rather than a single segmentation, deferring the choice of single grouping to later. It would be better to move that section to the related work}.
%
Rather than engineering the complexity and grouping criteria, we wish to learn them from data.\footnote{Of course, every unsupervised inference method requires inductive biases. Ours stem naturally from supervision in the meta-training set and density in the inferred clusters. 
% \updateblueafter{we removed the meta-clustering concept but still kept the meta-training concept as defined in line 67.}
% \note{since you have removed the characterization of "learning to cluster" as "meta-training" elsewhere in the paper, it should be removed from here too}
}  %\note{the previous sentence, if supported, is a strong novelty statement, but I am not sure it is accurate: It is true that the training set effectively "teaches" the model selection, so we do not need to specify it explicitly. Is it also true that we do not manually specify it? I am looking for the loss function, which formalizes the (meta)-learning criterion? It is  unclear to me whether  sect. 2.2 supports this claim. The loss is edge classification error in the trainin set (9), which is "true" as defined by ground truth. But I do not understand (10), or the effect of the condition $d_i \le d_j$ in (9). Is this only for complexity reasons? Or is that an explicit inductive bias? $d$ connects to (4), what is the inductive principle there? Can someone articulate clearly, in one sentence, IF TRUE, how the method chosen does not impose a separate and arbitrary grouping criterion, but instead ``meta-transfers'' the grouping criterion expressed by the ground truth in the training set, onto the test set? If this is not true, or if we cannot articulate it clearly, we will have to change this sentence.} \yifan{(9) and $d_i \le d_j$ indeed refers to an inductive bias. The idea is that nodes with low density tend to be those ones that have its neighborhood belonging to other classes or nodes that are on the boundary among multiple classes, thus connections to these nodes often result in edges connecting two nodes of different classes, which is undesirable and thus we establish connections with criteria $d_i \le d_j$. How we ``meta-transfer'': we learn the criterion for establishing edge connections (which are then used for determining clusters), through the supervision on the edge linkage and the relative order of node densities, which are both defined using ground-truth labels, but only on the ``meta-training'' set. And we infer these attributes, iteratively through a hierarchical process whose convergence is determined by the ``meta-training'' ground-truth as well, on unseen test-data whose ground-truth labels and the optimal clustering are unknown. Rather, we utilize the group criterion learnt on the ``meta-trainig'' set to make a best-possible estimation of the optimal clustering on the test-set. } 
%\note{I still do not see a clear answer to the question of whether the first statement of this paragraph is defensible. Right now that claim is in the abstract and in the second paragraph of the intro. If it is not defensible it must be removed. If it is dubious, it must be de-emphasized}
Clearly, this is not the data we wish to cluster, for we do not have any annotations for them. Instead, it is a different set of training data, the {\em meta-training} set, 
% \note{since reference to 'learning to cluster' as 'meta-training' has been removed elsewhere, it should be removed here too} 
% \updateblueafter{we removed the meta-clustering definition but kept the meta-training reference by defining it here.}
for which cluster labels are given, corresponding to identities that are disjoint from those expected in the test set.
For example, the test set might be an untagged collection of photos by a particular user, for which there exists a true set of discrete identities that we wish to discover, say their family members. While those family members have never been seen before, the system has access to different photo collections, tagged with different identities, during training. Our goal is to leverage the latter labeled training set to {\em learn how to cluster} different test sets with unknown numbers of different identities. This is closely related to ``open-set'' or ``open universe'' classification \cite{towardopenset, liu2017sphereface}.

% Such ``meta-clustering'' \cite{caruana2006meta}\david{This reference is not the same as what we are doing.}
% \begin{figure}[t]
%     \centering
%     \includegraphics[width=0.9\linewidth]{./figures/placeholder}
%     \caption{Teaser}
%     \label{fig:teaser}
% \end{figure}

We present the first hierarchical/agglomerative clustering method using Graph Neural Networks (GNNs). GNNs are a natural tool for learning how to cluster \cite{L-GCN, GCN-DS, GCN-VE}, as they provide a way of predicting the connectivity of a graph using training data. In our case, the graph describes the connectivity among test data, with connected components ultimately determining the clusters.

Our hierarchical GNN uses a novel approach to merge connected components predicted at each level of the hierarchy to form a new graph at the next level.  We employ a GNN to predict connectivity at each level, and iterate until convergence. While in unsupervised agglomerative clustering convergence occurs when all clusters are merged to a single node \cite{slonim1999agglomerative, quickshift}, or when an arbitrary threshold of an arbitrary model complexity criterion is reached, in our case convergence is driven by the training set, and occurs when no more edges are added to the graph by the GNN. There is no need to define an arbitrary model selection criterion. Instead, the ``natural granularity'' of the clustering process is determined  inductively, by the ground truth in the training set.
\begin{figure*}[t]
    \centering
    \includegraphics[width=0.95\linewidth]{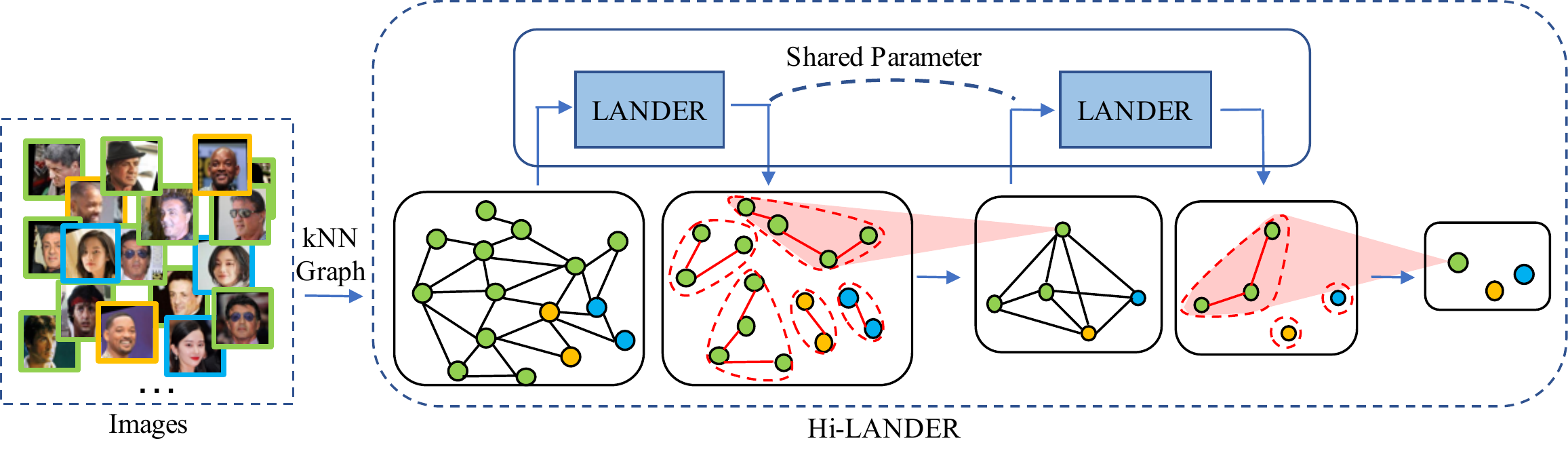}
    \caption{The proposed hierarchical clustering framework \HNAME. Images are embedded into a $k$-NN graph with their visual features. The green, yellow and blue image boundaries illustrate that they belong to three different classes; same for the color of the graph nodes. At each level of the hierarchy, our base \NAME model outputs a set of selected edges and thus intermediate clusters. These clusters are then grouped into super-nodes as input for the next level. The process continues til convergence, {\em i.e.}, when no more edges are added to the graph. Weights of \NAME are shared across multiple levels of the hierarchy. Best viewed in color.}
    % \NAME  first passes it through a GNN module with multiple stacked GAT layers. Then, $\{\hat{e}_{ij}\}$ are obtained for each edge, as the dense arrows are edges with higher probabilities while the dashed arrows are those with lower. Next, the density $\{\hat{d}_i\}$ is computed for each node, where darker nodes have higher density. After \NAME, the decoder finds edges in $E^\prime$ as the red arrows, and merges the resulting connected components into nodes for the next hierarchy.
    \label{fig:overall_framework}
\end{figure*}
Unlike prior clustering work using GNNs \cite{L-GCN, GCN-DS, GCN-VE}, we perform \emph{full-graph} inference to \emph{jointly} predict two attributes: linkage probabilities at the edges, and densities at the nodes, defined as the \updateblue{proportion of similar vertices that share the same label within a node's neighborhood \cite{DB-SCAN, OPTICS, GCN-VE}.}
% \tong{This is not accurate, probably some loose statement like "defined as the proportion of identically labeled nodes in one node's neighborhood"?} 
The densities establish a relative order between nodes \cite{OPTICS, GCN-VE}, which is then used to guide the connectivity. 
%\note{I am removing the following because "noisy" is not defined (what noise?), and there are no "constraints" per se, , just a relative order. Please correct if I am wrong. DELETE: helps avoid noisy edge connections by constraining the valid connections  to be from nodes with low-density to nodes of high-density. } 
% \yifan{agree, commenting off}  
Nodes at  the boundary between two ground-truth clusters, or nodes having a majority of their neighbors belonging to different classes, tend to have a low density, and accordingly a low expectation of linkage probability to their neighbors. 
Prior methods predict the edge connectivity as a node attribute on numerous sampled sub-graphs \cite{L-GCN, GCN-VE}; ours directly infers the full graph and predicts connectivity as an attribute of the edges. Also, prior methods require separate models for the two attributes of linkage probabilities and node densities, whereas ours infers them jointly. This is beneficial as there is strong correlation between the two attributes, defined by the ground truth. 
%\note{please check the accuracy of the previous sentence. REMOVING THE FOLLOWING benefits the learning because the node density and edge linkage share common ground [what is common ground?] as both are defined under a notion of similarity [what notion of similarity?]} 
% \yifan{prior sentence is accurate, commenting off the part to REMOVE}
A joint model also achieves superior efficiency, which enables hierarchical inference that would otherwise be intractable. Compared to the two separate models, we achieve a seven-fold speedup from 256s to 36s as shown in ~\mytabref{tbl:ablation_gat}.

\updateblue{In terms of accuracy, Our method achieves an average 54\% improvement in F-score, from 0.391 to 0.604, and an average 8\% increase in NMI, from 0.778 to 0.842 compared to state-of-art GNN based clustering methods \cite{GCN-VE, L-GCN} over the face and species clustering benchmarks as shown in ~\mytabref{tbl:diff_domain_train_test}}. 
% \note{I am looking at table 3 and I DO NOT SEE a 300\% improvement on the state of the art. From what number to what number of what score is this 300\%? If some methods have F-score of 0.6, a seven-fold improvement would be 1.8, clearly not possible. PErhaps you mean seven-fold error reduction? Or 30\%? Please check VERY CAREFULLY these statements and ensure that tehy do NOT engender confusion in the reader. Perhaps bringing Figure 2 here could be helpful? I find it the easiest figure to interpret.}
Furthermore, the pseudo-labels generated by our clustering of unlabeled data can be used as a regularization mechanism 
% \tong{this footnote seems too detailed to be included in the introduction, and is also difficult to understand (at least for me on the "Sigma Algebra"). I suggest we either move it to experiments or remove completely.} \note{I disagree. It is a footnote, which means that it is not necessary for the reader to understand it in order to proceed with the rest of the content, but it does serve the important purpose of signaling to potential reviewers who are more knowledgeable that we are not "self-training" naively. If you prefer to have this later, I have indicated a possible place in Section 4.5.} 
to reduce face verification error by 14\%, as shown in \mytabref{tbl:downstream_face}, from 0.187 to 0.159 in compared to state-of-art clustering methods, allowing us to approach the performance of fully supervised training at 0.136.

In the next section, we summarize our contributions in the context of prior related work. In \mysecref{sec:methodology} we introduce the technical innovations of our paper, and in \mysecref{sec:experiment} we detail our experiment results.  We discuss failure modes and limitations of our method in \mysecref{sec:discussion}.
%\note{describe succinct summary and roadmap for the paper}. 
% \updateblue{Our training and inference code are included in the supplementary material of the submission.}
% \note{Supplementary at submission? Or Github after acceptance?} \yifan{Yes, we plan to include as a code zip in supplementary at submission this time and we will make sure not to break any anonymity there. Github (DGL repo) open-sourcing will be a later separate item, after we finalize the arxiv paper.}
% \yifan{Yes, we plan to open-source them under the DGL's open-source eco-system upon finishing the review period together with putting it on arxiv. We are confirming on the publication approval ticket to make sure there are no issues before we do this.}

\section{Related Work and Contributions}

\noindent\textbf{Unsupervised Visual Clustering} \updateblue{Traditional unsupervised clustering algorithms utilize the notion of similarity between objects, such as K-means \cite{k-means} and hierarchical agglomerative methods \cite{newman2004fast, HAC, pons2005computing}.
% \note{Metric space. This sentence is wrong on many levels. First, arguably EVERY ML method operates in a metric space since, in the absence of a metric, you cannot define a probability measure, so you can do pretty much nothing. Second, if by operating on a metric space you mean 'distance-based' classifiers, such as those used in metric learning, then it is not true that trsaditional clustering are so. There are plenty of traditional methods that do not use a metric but instead use agglomeration scores. In any case, there is no such thing as a 'distance metric space'. A distance is a distance, and a metric is a metric. A metric is a bilinear form that is symmetric, positive semi-definite etc. for instance an inner product. A distance is a positive real-valued function that satisfies the triangular inequality. A metric space has a metric, from which you can construct a distance, but conversely a distance does not necessarily define a metric (it does in Euclidean space through the polarization identity). Use words precisely when describing technical concepts.}
% \note{please correct the grammar throughout. The article the does not go with plurals. These grammatical mistakes are unacceptable because there are easy ways of avoiding them} 
\cite{bonald2018hierarchical} extends Hierarchical Agglomerative Clustering (HAC) \cite{HAC} with a distance based on node pair sampling probability. 
% \note{no such thing as a 'distance metric'. It is like saying "today I ate an apple orange"}  
% Dated \note{dated??? What does "dated" means when referring to a method? Do methods have expiration dates?}  
Approaches based on persistent-homology \cite{persistent_homology} and singular perturbation theory \cite{ng2002spectral} deal with the scale-selection issue. \cite{DB-SCAN, OPTICS, H-DBSCAN} utilize a notion of density defined as the proportion of similar nodes within a neighborhood.
% defined as number of points within a certain neighborhood having a distance smaller than certain threshold, which are believed to be from the same ground-truth cluster.
% \note{what is a paradigm?}
% based on approximate nearest-neighbor.
% \note{one more time: distance metric does not exist} 
% \note{$l_1$ or $\ell_1$. These are distances, not metrics. A metric is a symmetric positive definite matrix.} 
Spectral clustering methods \cite{ng2002spectral, Spectral, von2007tutorial} approximate graph-cuts with a low-dimensional embedding of the affinity matrix via eigen-decomposition.
Graclus \cite{Graclus} provides an alternative to spectral clustering with multi-level weighted graph cuts. H-DBSCAN \cite{H-DBSCAN} removes the distance threshold tuning in \cite{DB-SCAN}. FINCH \cite{FINCH} proposes a first neighbor heuristic and generates a hierarchy of clusters. More recent unsupervised methods \cite{lin2018deep, lin2017proximity} utilize deep CNN features. \cite{zhu2011rank} proposes a Rank-Order distance measurement. Our hierarchical design relates the most to \cite{FINCH}, however, instead of the heuristic to link the first-neighbor of each node for edge selection, which is prone to error and has limited capability in dealing with large-scale complex cluster structures, we use a learnable GNN model.}
% \note{previous sentence is too dismissive. What is this heuristic? How is it bad?}

\noindent\textbf{Supervised Visual Clustering}
Supervised graph neural network-based approaches~\cite{L-GCN, GCN-DS, CDP, GCN-DS, GCN-VE} perform clustering on a $k$-NN graph. \updateblue{In contrast to these methods that produce only a single partition, our method generates a hierarchy of cluster partitions and deals with unseen complex cluster structures with a learnt convergence criterion from the natural granularity of the ``meta-training'' set. 
% \tong{do we want to avoid 'meta'?} 
In contrast to \cite{GCN-VE} which requires two separate models to perform edge connectivity and node density estimation, our method jointly predicts these two quantities with a single model of higher accuracy and efficiency (\mytabref{tbl:ablation_gat}). Furthermore, \cite{L-GCN, GCN-VE} estimate linkage as a node attribute on sub-sampled graphs, whereas we estimate it as an edge attribute with natural parallelization through full-graph inference and significantly reduce runtime (\mytabref{tbl:all_runtime}). \cite{nyupaper} uses a two-step process that first refines visual embeddings with a GNN and then runs a top-down divisive clustering, with testing limited to small datasets. In contrast, our method performs clustering as a graph edge selection procedure.}
% footnote{This is a master thesis that is not peer-reviewed and does not involve any experiments on large-scale datasets with complex cluster structures.}
% \note{weird place to introduce the name. Also, this needs more color as it undermines the narrative in the introduction. Here, it seems that the only difference is joint vs. two separate models}

% HAC \cite{HAC} uses a bottom-up approach to form clusters hierarchically with distance metrics. \cite{blondel2008fast} adopts the self-similar nature of social community networks and propose to iteratively detect a hierarchical structure. We follow the hierarchical design principal from \cite{HAC}. We adapt the idea of self-similarity from \cite{blondel2008fast} to visual clustering, where clusters of node features can be treated as input nodes in the next level of a hierarchy recurrently. But we operate on a GNN at each level instead of using traditional distance measurements.
\noindent\textbf{Hierarchical Representation}
 Hierarchical structures have also been extensively studied in many visual recognition tasks \cite{h_vision_language_task, h_future_action, h_mid_top, h_finegrain, h_track,h_2D_3D_match, hierarchical-VRD, fpn_object}. \updateblue{In this paper, our hierarchy is formed by multiple $k$-NN graphs recurrently built with clustering and node aggregation, which are learnt from the meta-training set.}
%  \updateblueafter{meta-training is still defined in the intro but meta-clustering as been removed throughout.}
%  \note{remove meta-training throughout} 
%  \updateblue{In \cite{hierarchical-VRD}, a hand-designed heterogeneous hierarchy of two graphs is constructed. Whereas, in ours, neither the number of hierarchies nor the node groupings are predefined, rather, they are learnt from the meta-training set.}
%  In \cite{hierarchical-VRD}, a heterogeneous hierarchy of two graphs is constructed to attend to different factors, which differs \note{differs how?} from our recurrently built hierarchy with arbitrary number of layers.
 Hierarchical representation has also been explored in the graph representation learning literature \cite{ying2018hierarchical, chen2017harp, bianchi2019hierarchical, huang2019attpool, hu2019hierarchical, lipov2020multiscale}. \updateblueafter{There, the focus is to learn a stronger feature representation to classify graph \cite{ying2018hierarchical} or input nodes \cite{hu2019hierarchical} into a closed set of class labels. Whereas, our goal is to ``learn'' to cluster from a meta-training set whose classes are disjoint to those of test-time.} 
 
%  There, the task focuses on graph \cite{ying2018hierarchical} or node classification \cite{hu2019hierarchical}, rather than image clustering, which is our goal. \note{this needs more color, as one can easily frame clustering as node classification}

\noindent\textbf{Graph Neural Networks in Visual Understanding}
\updateblue{The expressive power of GNNs in dealing with complex graph structures is shown to benefit many visual learning tasks
% \note{this opening is a bit superfluous. The list of references is important, as a sure way of getting your paper rejected is failure to cite work by a reviewer that the same reviewer believes related. It would be best to weave the references in but if this is a space-efficient way of llumping together loosely related work, that is ok. If instead these referencs are already discussed later, just remove the top part as redundant.}
% \yifan{these are loosely related works}
\cite{kipf2016semi, Graph-SAGE, FASTGCN, EdgeConv, GAT, defferrard2016convolutional, yan2019convolutional, yan2018spatial, chen2019graph, braso2020learning, weng2020gnn3dmot}.} \cite{Graph-SAGE} samples and aggregates embeddings of neighboring nodes. \cite{GAT} further advances \cite{Graph-SAGE} with additive attention. \cite{FASTGCN} uses a batch training scheme based on \cite{Graph-SAGE} to reduce computational cost. \updateblue{\cite{EdgeConv} performs node classification with edge convolution and feature aggregation through max-pooling. Our method differs from \cite{EdgeConv} in that we use a unified model that jointly learns node densities and edge linkages with two supervision signals.
% Both are defined using ground-truth labels of the meta-training set.
% \note{This undermines the introduction where we say that we do not state arbitrary grouping criteria but instead derive them from the ground truth classification. How do you reconcile?} 
Furthermore, our GNN learns the edge selection and convergence criteria for a hierarchical agglomerative process.}
% \note{intermediate? How about the non-intermediate ones? This is the first time this concept appears, and it comes out of the blue} 

\noindent\textbf{Contributions} We propose the first hierarchical structure in GNN-based clustering. Our method, partly inspired by \cite{FINCH}, 
% \note{better say "is inspired by X, but differs in Y"}
refines the graph into super-nodes formed by sub-clusters and recurrently runs the clustering on the super-node graphs, \updateblueafter{but differs in that we use a learnt GNN to predict sub-clusters at each recurrent step instead of an arbitrary manual grouping criterion.} At convergence, \updateblue{we trace back the predicted cluster labels on the super-nodes} 
% \note{we trace back what?}
from the top-level graph to the original data points to obtain the final cluster. 
% \note{this paragraph is more of a description of the method than a statement of the contribution}

Our method converges to a cluster based on the level of granularity established by ground truth labels in the training set. Although the identities are different from the test set, they are sufficient to implicitly define a complexity criterion for the clustering at inference time, without the need for a separate model selection criterion. 

\updateblue{To run multiple iterations of the GNN model efficiently and effectively,  we design a base model that approximates label-aware linkage probabilities and densities of similar nodes that share the same label. The densities are useful for additional regularization and refining edge selection. We refer to this base model as our \emph{Link Approximation aNd Density Estimation Refinement} (\NAME) module. Finally, we denote our hierarchical clustering method \HNAME and ~\myfigref{fig:overall_framework} illustrates its structure.}

The key innovation of our method is two-fold:  1) we produce a hierarchy of cluster partitions instead of a single flat partition of \cite{GCN-DS, L-GCN, GCN-VE}; 2) We perform full-graph inference to jointly predict attributes of both nodes and edges, whereas prior GNN methods used sub-graph inference and separate models for node and edge attribute prediction.
% These innovations are collectively responsible for improving image clustering performance with up to 300\% F-score boost and 23\% NMI increase over existing GNN based methods.
\updateblue{These innovations are collectively responsible for improving the clustering performance by an average of 54\% F-score and 8\% NMI  over existing GNN-based methods.
% \note{succinct summary of key experimental improvements}. 
 }
% \note{MUST Have a discussion of limitation and examples of failure modes, also ensure the experiments include computational cost and complexity}

\section{Methodology \label{sec:methodology}}

\subsection{\updateblue{Clustering with a $k$-NN Graph}}\label{sec:clustering_knn}
Formally, given a set of $N$ images $D = \{I_{i}\}_{i=1}^N$ and their corresponding visual embeddings $F = \{f_{i}\}_{i=1}^N$, we first construct an affinity graph $G = \{V, E\}$, where $|V|=N$, via $k$-nearest neighbors determined w.r.t. cosine similarity, {\em i.e.}, the inner-product of the normalized visual embeddings. Each image (for example one face crop) entails one object to cluster and represents a node in the graph, with the node feature being its visual embedding $f_i$.
% and \note{missing a subject}
% \note{relative to what distance or kernel?}
%Each image entails one object to cluster, for example, one face crop,
% \note{full image? Or local features/objects in the image?}
%and is represented as a node in the graph with the node feature being its visual embedding $f_i$. 
The edges connect each node to its $k$ neighbors. Per the clustering paradigms in \cite{DB-SCAN, H-DBSCAN, OPTICS, FINCH, L-GCN, GCN-VE, ARO}, a function $\phi$ takes as input the affinity graph $G$ and the node features $F$, and produces an edge subset $E^\prime \subset E$, {\em i.e.} $E^\prime = \phi(G, F)$. The resulting graph $G'=\{V, E^\prime\}$ is then split into connected components, with each corresponding to a cluster of nodes. Our method is built upon this $k$-NN graph based clustering paradigm.

\subsection{\updateblue{Hierarchical Generalization to \HNAME}}
%\yifan{Took out the ``... can be significantly dependent on $k$'' to avoid putting too much emphasize on the $k$ since it's a hyper-param that can be tuned with training/validation set.} \tong{I feel we still need something to motivate this hierarchical generalization, or in other words, what benefit can this generalization bring?} \yifan{what about ``in order to model the natural level of granularity of clusters in a dataset, we make a hierarchical generalization to the above single-level $k$NN based clustering paradigm. ''. Or maybe we simply omit it here, since here we focus on methodology and the motivation has already been stated in section1 and 2.} \tong{A recap doesn't harm our focus. Besides we do have such motivation in the beginning of section 3.3. }
% \updatered{The quality of the above procedure can be significantly dependent on $k$. A large $k$ is computationally infeasible, while a small $k$ might generate too many connected components. Here we introduce a hierarchical generalization, named \HNAME, to improve the result with a small $k$.} 
\updatered{In order to model the natural level of granularity of clusters in a dataset, we propose a hierarchical generalization to the above single-level $k$-NN based clustering paradigm.}
Given a set of initial visual embeddings $F$ and a small fixed value of $k$,\footnote{\updateblue{We emphasize that $k$ is a hyper-parameter tuned with the meta-training / validation set. 
% \updateblueafter{we still keep the meta-training definition but remove the meta-clustering concept throughout.}
% \note{see earlier note on meta-training}  
}}
% rather it is kept at a small fixed value across \textit{all} experiments. This is the case even for tests with drastically different cluster distributions, which is possible because our learnt GNN model will capture the natural granularity of different clustering processes via the hierarchical multi-level design (~\mytabref{tbl:diff_domain_train_test})}}\tong{This footnote seems to appear too early. More suitable as a part in the ablation/discussion sections.
% We further show in ~\myfigref{fig:k_sensitivity} that the method is robust to the choice of $k$, 
% \note{This potentially undermines the introduction unless you have an experiment that shows that the clustering is not affected by (or is robust to) the choice of $k$, other than for the computational complexity. If you herald automatic model selection without an explicit complexity term, well, $k$ is an arbitrary complexity term. For the pitch to work, here you would like to say something like: ADD: The choice of $k$ affects the computational cost of the procedure, but otherwise the final set of clusters is robust to the choice of $k$ as shown in fig. X, or table Z.} \yifan{yes, we have an experiment on $k$ robustness in \mytabref{tbl:test_k_robustness}} \note{then add a footnote of the kind: Although $k$ is an arbitrary complexity parameter, we show in Fig X that the method is robust to the choice of $k$.} 
\updatered{we iteratively generate of a sequence of graphs $G_l = \{V_l, E_l\}$ and the corresponding node features $H_{l} = \{h_i\}$, where $i = 1 \dots |V_l|$ and $l = 1 \dots $, using a base cluster function $\phi$ and an aggregation function $\psi$. Algorithm~\ref{algo:hierarchygraph} summarizes the proposed hierarchical generalization process.

To start, we define $G_1$ as the $G$ in \mysecref{sec:clustering_knn} and $H_1 = \{f_i\}$. \updateblue{
% The function $\phi$ is defined via 
% \note{the following equation does not define $\phi$. It would, implicitly, if $E$ was specified, but it is not.}
\updateblueafter{The function $\phi$ performs the following operation} 
\begin{equation}
    E^\prime_{l} = \phi(G_l, H_l),
\end{equation}
 taking as input the node features and $k$-NN graph at level $l$ and producing the selected edge subsets $E^\prime_{l}$. As a result, the graph $G_l^\prime = \{V_l, E_l^\prime\}$ is split into multiple connected components. We define the set of connected components in $G_l^\prime$ as $\{c^{(l)}_i\}_{i=1}^{|V_{l+1}|}$, where $c^{(l)}_i$ is the $i$-th element. }

In order to generate $G_{l+1}$, we obtain $V_{l+1}$, $H_{l+1}$ and $E_{l+1}$ as follows. First, we define the $i$-th node in $G_{l+1}$, $v_i^{(l+1)}$, as an entity representing the connected component $c_i^{(l)}$. \updateblue{Next, we generate the new node feature vectors through an aggregation function $\psi$, \updateblueafter{which performs}
\begin{equation}
    H_{l+1} = \psi(H_l, G^\prime_l),
    \label{eq:psi}
\end{equation}
}It aggregates the node features in each connected component $c_i^{(l)}$ into a single feature vector respectively. Finally, we obtain $E_{l+1}$ by searching for $k$-nearest-neighbors on $H_{l+1}$ and connecting each node to its $k$ neighbors.

The generation converges when no more new edges are added, {\em i.e.}, $E_l^\prime = \varnothing$. We define $L$ to be the length of the converged sequence. 
For the final cluster assignment, starting with $G_L$, we assign cluster identity (ID) $i$ to the connected component $c^{(L)}_i$, which propagates the ID $i$ to all its nodes $\{v^{(L)}_j|v^{(L)}_j\in c^{(L)}_i\}$. Then, each $v^{(L)}_i$ propagates its label to the corresponding connected component $c^{(L-1)}_i$ of the previous iteration. This ID propagation process eventually assigns a cluster ID to every node in $V_1$, and this assignment is used as the final predicted clustering. }

\if{false}
    %  \note{need to specify what the features are, lest the pitch in the intro is undermined -- ADD: At the lowest level, the features are simply $f_i$; and features at higher levels are learned through the agglomerative process. (or something like this)}
    $G_l$ denotes the $k$-NN graph and $G^\prime_l$ is the cluster partition of $G_l$ for level $l$. $V_l = \{v_j^{(l)}\}_{j=1}^{N_l}$ is the set of $N_l$ nodes at level $l$. We further define the $i$-th connected component inside $G^\prime_l$ as ${C^{(l)}_i}$ for $i=1...N_{l+1}$, each of which will serve as an intermediate cluster. We then begin with $H_0 = F$ and $G^\prime_0=(V_0, E^\prime_0) = (V,\varnothing)$, meaning that at the first level each visual embedding is treated as its own cluster.
     
    %  the node feature set and cluster partition at the first level, where $|V_0| = N_0$ and $E^\prime_0 = \varnothing$. This means that at the first level, each node is in its own cluster. 
    
    \HNAME then generates a sequence of graphs $\{G_l = (V_l, E_l)\}$ and intermediate cluster partitions $\{G^\prime_l = (V_l, E^\prime_l)\}$ through two functions, the base cluster function $\phi$ and aggregation function $\psi$. Analogous to previous single-level $k$-NN-based methods, the former is a base cluster function that takes as input the node features and $k$-NN graph at level $l$ and produces the edge subsets for that level $E^\prime_{l}$ via
    \begin{equation}
        E^\prime_{l} = \phi(G_l, H_l).
    \end{equation}
    In contrast, the latter is an aggregation function that takes as input the intermediate clusters $G^\prime_l = \{C^{(l)}_i\}_{i=1}^{N_{l+1}}$ and the node feature set $H_l$ at level $l$ to generate the node feature set for the next level $l+1$ via
    \begin{equation}
        H_{l+1} = \psi(H_l, G^\prime_l)
        \label{eq:psi}
    \end{equation}

    Note that $\psi$ aggregates the features of nodes within the same intermediate cluster $C^{(l)}_i$. After aggregation, we group the nodes $\{ v_j^{(l)} | v_j^{(l)} \in C^{(l)}_i \}$ to become the super-nodes $v_i^{(l+1)}$ at the next level and assign them the aggregated features $H_{l+1}$. \updateblue{When $E^\prime_l$ is empty, e.g., at the first level, $\psi$ reduces to the identity function, namely, $H_{0} = \psi(H_{0}, (V_0, \varnothing))$.} 
    % \david{The output of $\psi$ is a new set of embeddings, not a graph.  Also, the input argument order seems backwards.}
    % \david{What about defining this operation as a function that produces the updated graph for the next level?} \yifan{a caveat on that will be, then the output will be two items, one for $H_{l+1}$ and $G_{l+1}$ and it sounds more like an operation instead of a function.}

    Additionally, $\phi$ and $\psi$ are shared across the multiple levels of the hierarchy. During inference, these two function are applied iteratively to obtain the sequence of graphs ${G_l}$ and \updateblue{intermediate cluster partitions $G^\prime_l$, $l=1,2,...$. The process terminates when no more new edges are added, 
    % namely, when there is no difference between $G_l$ and $G_{l-1}$ 
    and $L$ is the level at which convergence occurs}.
    % \david{Is there any need to define $L$ prior to this point?  For example, just to avoid confusion, could we not just use $l = 1,2,\dots$ in previous instances and then postpone the definition of $L$ to here as the value at convergence?} 
    $L$ varies from dataset to dataset. The asymptotic granularity is not explicitly informed, for instance by choosing a maximum number of clusters, or through an explicit model selection criterion, but is instead informed by the meta-training set. \note{ditto} In fact, different ground-truth classes even within the same test-set converge at different levels, specified by the natural granularity of the classes (Figure 3 in the appendix).
    % \note{must say something about terminating condition. For instance: The process terminates when edges are no longer updated. The asymptotic granularity is not explicitly informed, for instance by choosing a maximum number of clusters, or through an explicit model selection criterion, but is instead informed by the training set. -- NOTE: it would be good to show an experiment where different annotaions (e.g. where labels correspond to families, vs. individuals) produce clustering methods that settle at different granularity} \yifan{we have a plot ~\myfigref{fig:hannah_deeper} that shows different test ground-truth classes converge at different levels in the hierarchy. In faces training/testing, classes are flat, so there will be personA class and personB class, but no personA class and familyA class. But good to check if there are non-flat / nested classes in species.}  
    For the final cluster assignment, starting with $G_L$, we assign cluster ID $i$ to intermediate cluster $C^{(L)}_i$, and $C^{(L)}_i$ propagates the cluster ID $i$ to all its nodes $\{v^{(L)}_j|v^{(L)}_j\in C^{(L)}_i\}$. Naturally, each $v^{(L)}_i$ propagates its label to the corresponding intermediate cluster $C^{(L-1)}_i$ in the previous hierarchy. This ID propagation process will eventually assign a cluster ID to every node in \updateblue{$V_0$},
    % \david{I thought $|V_0| = |V| = N$, in which case this expression is awkward}
    and this assignment is used as the predicted clustering. 
\fi

% \updatered{
% During this process, the two functions $\phi$ and $\psi$ are applied iteratively to obtain the sequence of $\{G_l\}$.}
% The inference on unseen datasets can result in sequences with arbitrary length , thus we propose to share the $\phi$ and $\psi$ during the process.
% \david{The way this section is framed could create the impression that Algorithm 1 is merely an existing approach that we are borrowing to create \HNAME by filling in some components. In this regard, could not Section 2.1 and Algorithm 1 be reframed as something like "High-Level Design of \HNAME" or related?  Or alternatively, could we break Section 2.1 into two subsections, where the first is something like "Clustering with a $k$-NN Graph" (not our contribution) and the second is called "Hierarchical Generalization to Hi-LANDER" (our contribution).} 
\updateblue{In the following sections, we describe the design of the base cluster function $\phi$, the aggregation function $\psi$ and how we learn the overall \HNAME model with a meta-training set. We also adopt the name \NAME to refer to our underlying single-level model, akin to a single iteration of \HNAME. }
% \david{But the section headings themselves do not indicate how they relate to these functions.  And actually Section 4.2 and 4.3 contain more details beyond just the descriptions of these functions.  In this regard I'm wondering if these headings could be reworded a bit.} \note{+1}

\begin{algorithm}[]
\SetAlgoLined
 \label{algo:hierarchygraph}
 \textbf{Input} $N$, $F$, $k$\;
 \updatered{
 $l \gets 1$ \;
 $H_1 \gets F$ \;
 }
 %$G^\prime_0 \gets \{V_0, \varnothing\}$ \;
 %$H_0 \gets F$ \;
%  \For{$i \gets 1$ \textbf{to} $N$}{
%   $h^{(0)}_i \gets \bar{h}^{(0)}_i \gets f_i$ \;
%  }
 \While{not converged}{
%   $\{v^{(l)}_i\} \gets \{C^{(l)}_i\} \gets \text{Connected}(G^\prime_{l-1})$ \;
\updatered{
$G_l \gets k\text{-nearest-neighbor} ( H_{l}, k) $\;
  $E^\prime_l \gets \phi(G_l, H_l)$ \;
  $G^\prime_l \gets \text{connected-components}(E^\prime_l)$ \;
  $H_{l+1} \gets \psi(H_l, G^\prime_l) $ \; 
%   \david{But $C_l$ has not been used in Algorithm 1; should be $G^\prime_l$?}
  $l \gets l + 1$ \;
  }
%   $\{e^{(l)}_{ij}\}, \{d^{(l)}_i\} \gets \text{Encode}(G_l, \{[h^{(l)}_i, \bar{h}^{(l)}_i]\})$ \;
%   $G^\prime_l, E^\prime_l \gets \text{Decode}(\{e^{(l)}_{ij}\}, \{d^{(l)}_i\})$ \;
 }
 \updatered{
 $\text{ID} \gets \text{id-propagation}(\{G_l\}, \{G_l^\prime\})$ \;
 \textbf{Return} ID 
 }
 \caption{\updatered{Hierarchical Generalization}}

\end{algorithm}

\subsection{Realizing the Cluster Function $\phi$}
\updateblue{
% There are two desired properties of the base cluster function $\phi$. One is on clustering \emph{accuracy} and the other is on \emph{efficiency}. 

To achieve high accuracy, we design $\phi$ as a learnable GNN model for clustering in a supervised setting 
% \note{If we are comfortable with the pitch in the intro, this would be the place to remind the reader that we do not engineer the grouping criterion but learn it from the training set.} 
to deal with complex cluster structures, where each node $v_i$ in $V$ comes with a cluster label $y_i$, but only in the meta-training set. Unlike unsupervised clustering methods, we do not engineer an explicit grouping criterion but learn it from data.} State-of-art supervised clustering methods \cite{L-GCN, GCN-VE} show that density and linkage information are effective supervision signal to learn the GNN model and we use both of them. 
\updateblue{However, unlike prior work, to improve both efficiency and  accuracy, we jointly predict these two quantities using the embeddings produced by a single graph encoder. The linkage and density estimates are then passed through a graph decoding step for determining edge connectivity and thus cluster prediction. Below details our \NAME design.} 
% Below, we detail the design of our joint linkage and density estimation model - \NAME.
% With these considerations, we propose a unified joint linkage and density estimation model named \NAMETh, where it does a single full-graph inference to jointly predict the graph edge linkage and node density information, which are then passed through a graph decoding step for efficient and effective cluster prediction. \note{the above paragraph is repetitive; the same point has been made at least twice before}

% when presented with a $k$-NN graph $G$, outputs linkage and density estimators that are then passed through a graph decoding step for producing intermediate cluster assignments.

\noindent\textbf{Graph Encoding}
%\myfigref{fig:unit_model_illustration} illustrates our unit model architecture. 
%\myfigref{fig:overall_framework} includes \NAME architecture. 
% We adopt the Graph Attention Network (GAT) \cite{GAT} as the basis of our graph  encoder. % \updateblue{Compared to a vanilla graph convolutional layer (GCN), GAT adaptively weights each node's neighboring vertices in relevance to the prediction targets.}
% the ground-truth node and edge attribute values.}
% \note{what information? how is relevance measured? } 
\updatepurple{For each node $v_i$ with corresponding input feature $h_i$, a stack of Graph Attention Network (GAT) \cite{GAT} layers encode each $h_i$ as the new feature or embedding $h^\prime_i$. In general though, we found that alternative encoders (e.g., vanilla graph convolutional network layers), produce similar performance (see appendix)}.

% , e.g., GCN, Note that these encoding layers can also be swapped to any GCN architecture, including a vanilla GCN layer. % We show the sensitivity of our model to this design choice in ~\mytabref{tbl:ablation_gat}.

%\david{Isn't this material below and MLP just part of ``Graph Encoding"?}
\noindent\textbf{Joint prediction for density and linkage}
For each edge $(v_i, v_j)$ in $E$, we concatenate the source and destination node features obtained from the encoder as $[h^\prime_i, h^\prime_j]$, where $[\cdot, \cdot]$ is the concatenation operator, and feed it into a Multi Layer Perceptron (MLP) layer followed by a softmax transformation to produce the linkage probabilities $p_{ij} = P(y_i = y_j)$, 
% \note{Probability of an event is writtenas $P(\omega)$, not $Prob(\omega)$. } 
{\em i.e.}, an estimate of the probability that this edge is linking two nodes sharing the same label. 
We also use this value to predict a node pseudo-density estimate $\hat{d}_{i}$, which measures the similarity-weighted proportion of same-class nodes in its neighborhood.\footnote{Note that $\hat{d}_i$ is only a density proxy, not a strict non-negative density that sums to one.}

For this purpose, we first quantify the similarity $a_{ij}$ between nodes $v_i$ and $v_j$ as the inner product of their respective  node features, {\em i.e.}, $a_{ij}=\langle h_i, h_j\rangle$. 
% \note{use $\langle, \ \rangle$, not $<, \ >$. }
Subsequently, we compute corresponding edge coefficients as $\hat{e}_{ij}$ as
\begin{equation}
    \hat{e}_{ij} = P(y_i = y_j) - P(y_i \neq y_j)
\end{equation}
% \note{Prob vs. P as above}
where $j$ indexes the $k$ nearest neighbors of $v_i$.
We may then define $\hat{d}_{i}$  as 
\begin{equation} \label{eq:approximate_density}
    \hat{d}_{i}  = \frac{1}{k}\sum_{j=1}^{k} \hat{e}_{ij} \cdot a_{ij}.
\end{equation}
This estimator is designed to approximate the ground-truth pseudo-density $d_i$, which is obtained by simply replacing $\hat{e}_{ij}$ in Eq.~\ref{eq:approximate_density} with $e_{ij} = \mathbbm{1}(y_i = y_j) - \mathbbm{1}(y_i \neq y_j)$
% \note{Are these indicator functions? Earlier on $I$ denotes an image, or the data. Here you change notation?}  \yifan{yes, these are indicator functions} \note{then you need to change the notation. Either call the image data $x$, or call the indicator (characteristic) function $\chi$.} 
using the ground-truth class labels, where $\mathbbm{1}$ is the indicator function.  By construction, $d_i$ is large whenever the most similar neighbors have shared labels; otherwise, it is small. And importantly, by approximating $d_i$ in terms of $\hat{e}_{ij}$ via $p_{ij}$, the resulting joint prediction mechanism reduces parameters for the prediction head during training (see ~\mysecref{sec:learning_hilander} below), allowing the two tasks to benefit from one another.

\noindent\textbf{Graph Decoding} 
%\updatered{Frequently a clustering algorithm uses a decoding process to retrieve the cluster assignment\cite{OPTICS, GCN-VE}, and ours is adapt to the graph encoder.}  
Once we obtain the linkage probabilities and node density estimates, we convert them into final clusters via the following decoding process. \updatered{Prior methods rely on an analogous decoding step \cite{OPTICS, GCN-VE}; however, herein we tailor this process to incorporate our joint density and linkage estimates.}  Initially we start with $E^\prime = \varnothing$. Given $\hat{e}_{ij}$, $\hat{d}_{i}$, $p_{ij}$ and an edge connection threshold $p_\tau$,  we first define a \emph{candidate edge set $\mathcal{E}(i)$ } for node $v_i$ as
% \note{need to point to experiment on sensitivity of outcome to any parameters, including thresholds}
\begin{equation}
    \mathcal{E}(i) = \{j|(v_i, v_j)\in E \text{ and } \hat{d}_{i}\leq \hat{d}_{j}  \text{ and } p_{ij} \geq p_\tau\}.
\end{equation}
For any $i$, if $\mathcal{E}(i)$ is not empty, we pick
\begin{equation}
j = \operatorname*{argmax}_{j\in\mathcal{E}(i)} \hat{e}_{ij}
\end{equation}
and add $(v_i, v_j)$ to $E^\prime$. 
% \tianjun{added validation set here}
\updateblue{We emphasize that the selection of the edge connection threshold $p_\tau$ is a hyper-parameter tuning process ~\emph{only} on the validation set split from the meta-training set. It stays fixed after meta-training. This is different from the arbitrary parameter selection in unsupervised agglomerative clustering where the selection criteria will likely need to change across different test sets.}
% The condition $p_{ij} \geq p_\tau$ is an optional filter, and could be switched to other formula. One replacement is $\cos(f_i, f_j) \geq s_\tau$, where $s_\tau$ is a similarity cutoff threshold.

Additionally, the definition of 
% \note{$\cal E$ is not a definition. (5) is a definition. either "the definition of $\cal E$" or "the definition (5)".}
$\mathcal{E}(i)$ ensures that each node $v_i$ with a non-empty $\mathcal{E}(i)$ adds exactly one edge to $E^\prime$. On the other hand, each node with an empty $\mathcal{E}(i)$ becomes a peak node with no outgoing edges.
% \note{reword the previous sentence} 
\updateblue{Meanwhile, the condition $d_i \le d_j$ introduces an inductive bias in establishing connections. As nodes with low density tend to be those ones having a neighborhood that overlaps with other classes, or nodes on the boundary among multiple classes, connections to such nodes are often undesirable.}
% often result in edges connecting two nodes of different classes which are
%
After a full pass over every node, $E^\prime$ forms a set of connected components $G^\prime$, which serve as the designated clusters.
% \david{Should "Graph Decoding" be placed last?  Seems a bit strange to place between the base LANDER description and Learning LANDER.} \yifan{I think primarily this is to introduce the inductive bias based on density.}

\subsection{Realizing the Aggregation Function $\psi$ \label{sec:aggregation_function}}

Recall that we denote $c^{(l)}_i$ to be the $i$-th connected component in $G^\prime_l$. To build $G_{l+1}=\{V_{l+1}, E_{l+1}\}$, we first convert $c^{(l)}_i$ in $G_{l}$ to node $v^{(l+1)}_i$ in $V_{l+1}$. 
% \updategreen{Central to this conversion is keeping $V_{l+1}$ and $V_{l}$ in the same feature space such that shared model weights can be applied across all levels.} \yifan{remove since we already talk about that in 3.5}
We define two node feature vectors for the new node, namely the identity feature $\tilde{h}^{(l+1)}_i$ and the average feature $\bar{h}^{(l+1)}_i$ as
\begin{equation}
    \tilde{h}^{(l+1)}_i  = \tilde{h}^{(l)}_{m_i} ~~~\mbox{and}~~~ \bar{h}^{(l+1)}_i  = \frac{1}{|c^{(l)}_i|}\sum_{j\in c^{(l)}_i}   {\tilde{h}^{(l)}_{j}},
\label{eq:feature}
\end{equation}
where $m_i = \operatorname*{argmax}_{j \in c^{(l)}_i} \hat{d}^{(l)}_{j}$, represents the peak node index of the connected component $c^{(l)}_i$. Additionally, in the first level, $\tilde{h}^{(0)}_i = \bar{h}^{(0)}_i = f_i$, where $f_i$ is the visual embedding feature.
% \updategreen{How to construct the next-level input feature $h^{l+1}_i$ for the base cluster function $\phi$ of node $v^{(l+1)}_i$ from $\tilde{h}^{(l+1)}_i$ and $\bar{h}^{(l+1)}_i$ is a hyper-parameter to be decided. We can use one of them or combine them together.

\updateblue{The next-level input feature for the base cluster function $\phi$ of node $v^{(l+1)}_i$ is the concatenation of the peak feature and average feature, {\em i.e.}, $ h^{l+1}_i = [\tilde{h}^{(l+1)}_i, \bar{h}^{(l+1)}_i]$. We empirically found that directly using one of the features produces similar performances as the concatenation on some validation sets and we left this as a hyper-parameter.}
The identity feature $\tilde{h}^{(l)}_i$ can be used to identify similar nodes across hierarchies, while the average feature $\bar{h}^{(l)}_i$ provides an overview of the information for all nodes in the cluster. 
% \yifan{discuss the super-node feature concatenation as a hyper-parameter, defer to supplementary for further ablations.}
% The combination is done by concatenation, $ h^{l+1}_i = [\tilde{h}^{(l+1)}_i, \bar{h}^{(l+1)}_i]$.} 
%\updateblue{Then the next-level input feature for the base cluster function $\phi$ of node $v^{(l+1)}_i$ is the concatenation of the peak feature and average feature, {\em i.e.}, $ h^{l+1}_i = [\tilde{h}^{(l+1)}_i, \bar{h}^{(l+1)}_i]$.
%The identity feature is used for $k$-NN graph construction.} 

\subsection{\HNAME Learning\label{sec:learning_hilander}} 
% \david{This subsection title is too general since Hi-LANDER design choices are also described in sections 4.1 and 4.2.}\note{+1}
% In this section, we detail the various design choices inside \HNAME including the feature set aggregation function $\psi$, the early stopping criteria and \HNAME model training strategies. \note{perfunctory, not very informative, opening}
% In order to learn the hierarchical GNN clustering method end-end, we need to define the aggregation function $\psi$ to merge features on intermediate graph nodes and assign them as features for super nodes in the subsequent levels. 
% Note that,  are always in the input visual embedding feature space for any $l$. This design ensures that the same model parameters work with any $G_l$. 
Because the merged features for super nodes, $\tilde{h}^{(l+1)}_i$ and $\bar{h}^{(l+1)}_i$, always lie within the same visual embedding space as the node features $h^{(l)}$ of the previous level, the same GNN model parameters can be shared across multiple levels of the hierarchy structure in learning the natural granularity of the cluster distribution of the meta-training set.

\noindent\textbf{Hierarchical Training Strategy}
Given $k$ and the ground truth labels, we can determine the level $L$ at which the hierarchical aggolemeration convergences. \updateblue{Thus, we build the sequence of graphs $\{G_l\}$ with respect to the algorithm depicted in Algorithm ~\ref{algo:hierarchygraph}, the only difference being that we use the ground-truth edge connections $\{E^{\prime_{gt}}_l\}$ at all levels and thus ground-truth intermediate clusters $\{G^{\prime_{gt}}_l\}$ for graph constructions. We initialize \NAME, and train it on all intermediate graphs $\{G_l\}$.}
% \david{This part might not be clear from the details provided.  In particular, the sequence of graphs I believe are generated by the ground-truth linkage at all levels, not the output of \HNAME.  Hence the sequence of graphs is not exactly the product of Algorithm 1.} 
In one epoch, we loop through each $G_l$, perform a forward pass on graph $\{G_l\}$, compute the loss as will be defined next, and then update the model parameters with backpropagation. 

% Learining LANDER
% \noindent\textbf{Learning \NAME} 
\noindent\textbf{Training Loss}
The \HNAME model is trained using the composite loss function given by
\begin{equation}
\mathcal{L} = \mathcal{L}_{conn} + \mathcal{L}_{den}.
\end{equation}
% The first term provides supervision on  pair-wise (linkage prediction) and the second on the neighborhood (density estimation) level, \updateblue{both defined with the ground-truth cluster labels over the meta-training set, which are disjoint from the classes at test-time. During test-time, the ground truth labels are unknown and we use the \NAME module within \HNAME to best estimate these edge and node attributes at each level for the optimal clustering.}\david{The content below eq.~6 is somewhat redundant and could probably be cut to save space.  Note that the test-time is Algorithm 1, which already clearly defines how the linkage and densities are computed.  This section could be devoted just to training.}
% \note{reword the previous sentence to make it flow. It should be clear that these losses are computed on the trainint set only, and they use ground truth labels. They do not make sense on the test set.}
\updateblue{The first term $\mathcal{L}_{conn}$ provides supervision on  pair-wise linkage prediction via the average per-edge connectivity loss}
\begin{equation} \label{eq:connectivity_loss}
\mathcal{L}_{conn} = -\frac{1}{|E|} \sum_{(v_i,v_j)\in E} l_{ij},
\end{equation}
where $l_{ij}$ is the per-edge loss in the form
% \note{You cannot have two things called the same name. If you call this the loss (Which is appropriate), the average in (8) is the risk, or average loss. Normally the average is over data, here it is over vertices, so you can give it a different name if you think this may cause confusion.} 
\begin{equation}
\begin{small}
    l_{ij} = 
    \begin{cases}
        q_{ij}\log p_{ij} +(1 - q_{ij}) \log (1 - p_{ij}), & \text{if } d_i \leq d_j\\
        0,              & \text{otherwise}
    \end{cases}.
\end{small}
\end{equation}
\updateblue{
Here the ground truth label $q_{ij} =  \mathbbm{1}(y_i = y_j)$ indicates whether the two nodes connected by the edge belong to the same cluster, and can be computed across all levels as described previously (similarly for the ground-truth $d_i$ derived from the $q_{ij}$ values).}
% Note the condition $d_i \leq d_j$ above makes the connectivity predictor specialize to the way it is used in decoding. \note{unclear what the previous sentence means} That is, we only allow an edge to connect a source node to a neighbor that has a higher density. \note{this does not clarify it}
Meanwhile, the second term $\mathcal{L}_{den}$ represents the neighborhood density average loss given by
\begin{equation} \label{eq:density_loss}
    \mathcal{L}_{den} = \frac{1}{|V|} \sum_{i=1}^{|V|} || d_{i} - \hat{d}_{i} ||_{2}^{2}.
\end{equation}
During training, both $\mathcal{L}_{conn}$ and $\mathcal{L}_{den}$ are averaged across data from all levels. 
% This measures the difference between the density estimate and the corresponding ground-truth.\david{This sentence is not really needed}
Note that prior work has used conceptually related loss functions for training GNN-based encoders~\cite{GCN-VE}; however, ours is the only end-to-end framework to do so in a composite manner without introducing a separate network or additional parameters.
% \david{Should this be moved until after the loss functions are actually defined? Maybe the end of Section 2.2?} 

\section{Experimental Results\label{sec:experiment}}
\updateblue{We evaluate \HNAME across clustering benchmarks involving image faces, video faces, and natural species datasets. First, we show the sensitivity of our method to early-stopping and illustrate that it is only used to reduce complexity without affecting accuracy. We also illustrate ablation experiments over the model components. We then evaluate clustering performance under both settings of same train-test and unknown test-distributions. We further show the advantage of \HNAME via a semi-supervised face recognition task with pseudo label training. Finally, we analyze the runtime cost.}
We compare with the following baselines. \updateblue{The unsupervised methods include DB-SCAN \cite{DB-SCAN},  ARO \cite{ARO}, HAC \cite{HAC}, H-DBSCAN \cite{H-DBSCAN},  Graclus \cite{Graclus} and FINCH \cite{FINCH}, where the latter four are hierarchical baselines.}
% The unsupervised methods include Hierarchical Agglomerative Clustering (HAC) \cite{HAC}, Density-Based Spatial Clustering of Applications with Noise (DB-SCAN) \cite{DB-SCAN},  Approximate Rank Order Clustering (ARO) \cite{ARO}, Hierarchical DBSCAN (H-DBSCAN) \cite{H-DBSCAN},  Multi-level Weighted Graph-Cuts (Graclus) \cite{Graclus} and FINCH \cite{FINCH}. 
The supervised baselines include L-GCN \cite{L-GCN}, GCN-V \cite{GCN-VE} and GCN-E \cite{GCN-VE}. Hyperparameters for the baselines are tuned to report their best performances respectively. For example, we tune the optimal MinPts parameter for H-DBSCAN. Supervised GNN baselines have their best parameters tuned with the validation sets (part of the meta-training set), e.g., we tune the optimal $k$-NN $k$ and $\tau$ parameters for GCN-V/E.
% we use $knn-k$ as 80 and $\tau$ as 0.8 which are optimal parameters originally reported in their paper.

% To understand the performance improvement, we have a deep dive into how hierarchical framework works on dataset largely varying in cluster size.

\subsection{Evaluation Protocols}
\updateblue{\noindent\textbf{Datasets} For face clustering, we use the large-scale image dataset TrillionPairs \cite{TrillionPairs} and randomly choose one-tenth (660K faces) for training. 
For testing, we use IMDB (1.2 million faces) \cite{IMDB} and Hannah (200K faces) \cite{Hannah}. Hannah has no overlapping person identities with the TrillionPairs training set, whereas IMDB has a small overlap (less than $2\%$). Features for all face datasets are extracted from a state-of-the-art embedding model~\cite{arcface} trained on TrillionPairs. The average cluster size of TrillionPairs, IMDB, and Hannah are 37, 25 and 800 respectively.
% There exist additional face benchmarks, for example the ones from \cite{GCN-VE}; however, we cannot use them due to data sensitivity issues.
For species clustering, we use iNaturalist2018 \cite{van2018inaturalist}. We follow the open-set train-test split for image retrieval as in \cite{brown2020smooth}, where the training (320K instances) and testing (130K instances) classes are disjoint. Both splits have similar cluster size distributions with an average of 56 instances per class. Features are extracted from a ResNet50 pretrained object retrieval model from \cite{brown2020smooth}. Table 6 of the appendix shows detailed statistics of all datasets. For all clustering training sets, we reserve $20\%$ for validation and hyper-parameter tuning. When finalized, we re-train on the entire training split with fixed hyper-parameters.
% \noindent\textbf{Clustering Train-Test Split} For face clustering, Hannah has no overlapping person identities with the TrillionPairs training set, whereas IMDB has a small overlap (less than $2\%$). For species clustering, we follow the open-set train-test split for image retrieval tasks defined in \cite{brown2020smooth}, where the classes for training and testing are disjoint. Table~\ref{tbl:dataset_stats} provides detailed statistics for all the datasets used for the two settings of clustering with same train-test and unseen test distributions. For all clustering training sets, we reserve $20\%$ for validation and hyper-parameter tuning. When finalized, we re-train on the entire training split with fixed hyper-parameters.  % put inside introduction for datasets.
% \noindent\textbf{Pseudo-Label Learning Datasets} 
We use Deepglint and IMDB datasets for pseudo label training for face recognition and evaluate using the openset IJBC \cite{IJBC} benchmark. }
% For the image retrieval downstream task on species images, we use iNaturalist2018 \cite{van2018inaturalist} and its disjoint split \cite{brown2020smooth}.

% \begin{table}[tbh]
% \begin{center}
% \begin{footnotesize}
% % \setlength{\tabcolsep}{2pt}
% \centering
%     \begin{adjustbox}{max width=\linewidth}
%             \begin{tabular}{c|c |c | c}
%     	Dataset & Images & Entities & Mean Cluster Size \\
%     	\hline
%     	\hline
%     	TrillionPairs-Train \cite{TrillionPairs} & 669,560 & 18,084 & 37.0 \\ 
    	
%     	Hannah-Test \cite{Hannah} & 201,240 & 251 & 801.8 \\
    	
%     	IMDB-Test \cite{IMDB} & 1,265,173 & 50,289 & 25.2 \\
    	
%     	IMDB-Test-SameDist \cite{IMDB} & 614,002 & 18,084 & 34.0 \\
    	
%     	iNat2018-Train \cite{van2018inaturalist} & 324,418 & 5,690 & 57.0 \\
    	
%     	iNat2018-Train-DifferentDist \cite{van2018inaturalist} &  51,696 & 5,690 & 9.0 \\
    	
%     	iNat2018-Test \cite{van2018inaturalist} & 135,660 & 2,452 & 55.3 \\

%     	\hline
%     \end{tabular}
%     \end{adjustbox}
%     \caption{\label{tbl:dataset_stats} Statistics on Datasets \yifan{put in supplementary.}}
% \end{footnotesize}
% \end{center}
% \end{table}
%  \yifan{put inside respective tasks} 
% 
\noindent\textbf{Evaluation Metrics} \updateblue{For clustering, we report the Normalized Mutual Information (NMI) \cite{NMI} capturing both homogeneity and completeness. We also report the pairwise and bicubed F-score which are two types of harmonic mean of the precision and recall of clustering prediction, denoted by $F_{p}$ and $F_{b}$. We report the standard face recognition metrics, including 
% \note{as customary. You may want to say that these choices are standard and you are not just picking the diagnostics that favor your method} 
False Non Match Rate (FNMR) @ various False Match Rate (FMR) for verification and False Negative Identification Rate (FNIR) @ different False Positive Identification Rate (FPIR) for identification.}
% For species image retrieval, we report the Recall@K metric with $K$ = 1, 4, 16, 32 as in \cite{brown2020smooth}.} 

\subsection{Implementation Details}

\updateblue{We use the validation sets to choose our optimal meta-training hyper-parameters. $k$ is set to 10 for $k$-NN graph building and is fixed for inference for \textit{all} settings and test-sets. $p_\tau$ is set to 0.8 for face clustering and 0.1 for species. 
% We use the identity feature aggregation for face clustering while the concatenation of identity and average feature (detailed in ~\mysecref{sec:aggregation_function}) for species.
Both face and species clustering use the identity feature aggregation (detailed in ~\mysecref{sec:aggregation_function}).
All validation sets are part of the meta-training sets and we have no access to any test information during hyper-parameter tuning.} Due to space limitations, sensitivity analysis to these hyper-parameters and additional details are in the appendix.
% \note{pretty uninformative section; best to put there (some) most critical information rather than spending a paragraph to say to go and read the supplementary material}

\subsection{Ablation Experiments}
% \noindent\textbf{Early Stopping Criteria}

\begin{figure}[tbh]
    \begin{center}
    \centering
\begin{adjustbox}{max width=\linewidth}
        \includegraphics[width = 0.5\textwidth]{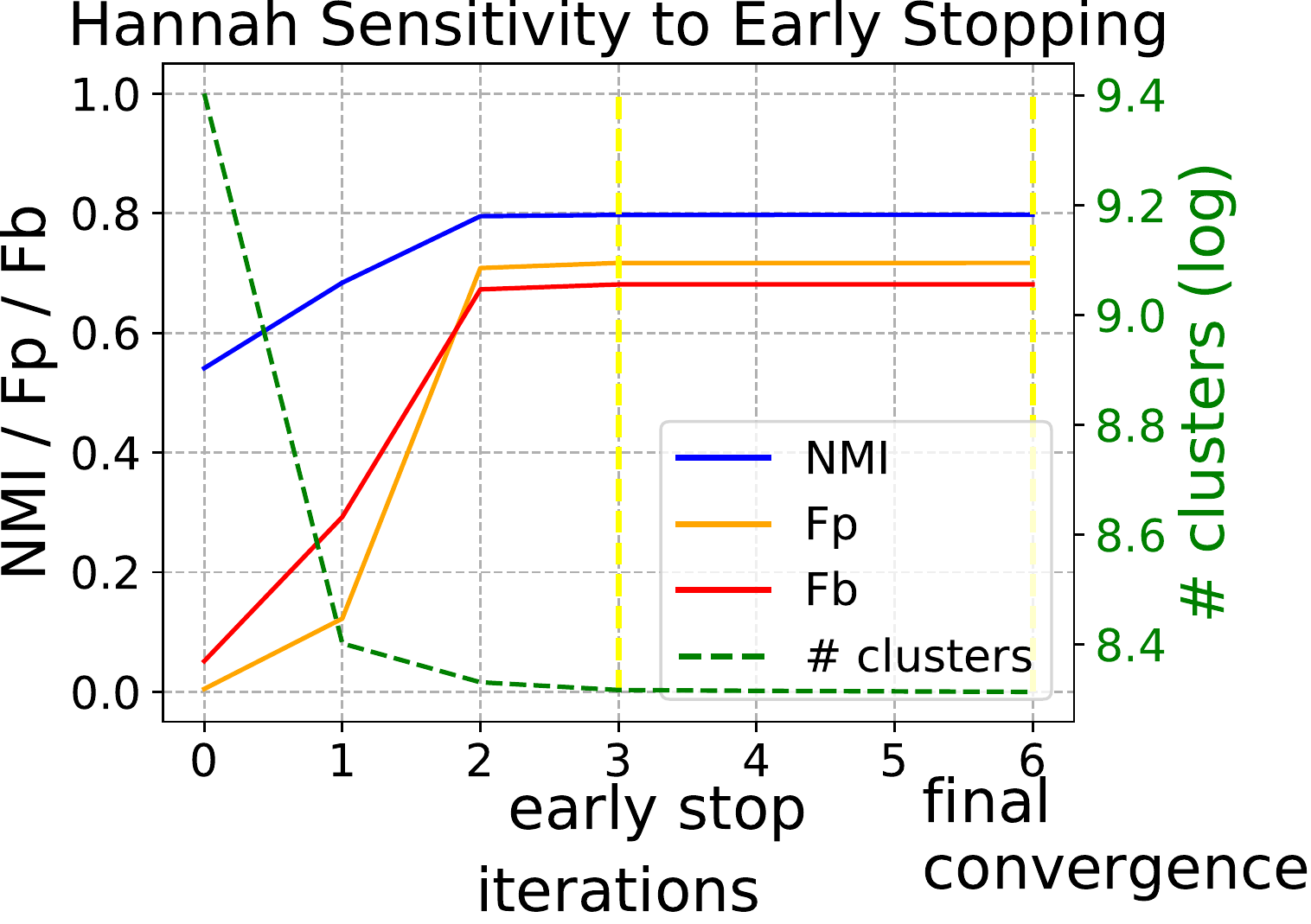}
        \includegraphics[width = 0.5\textwidth]{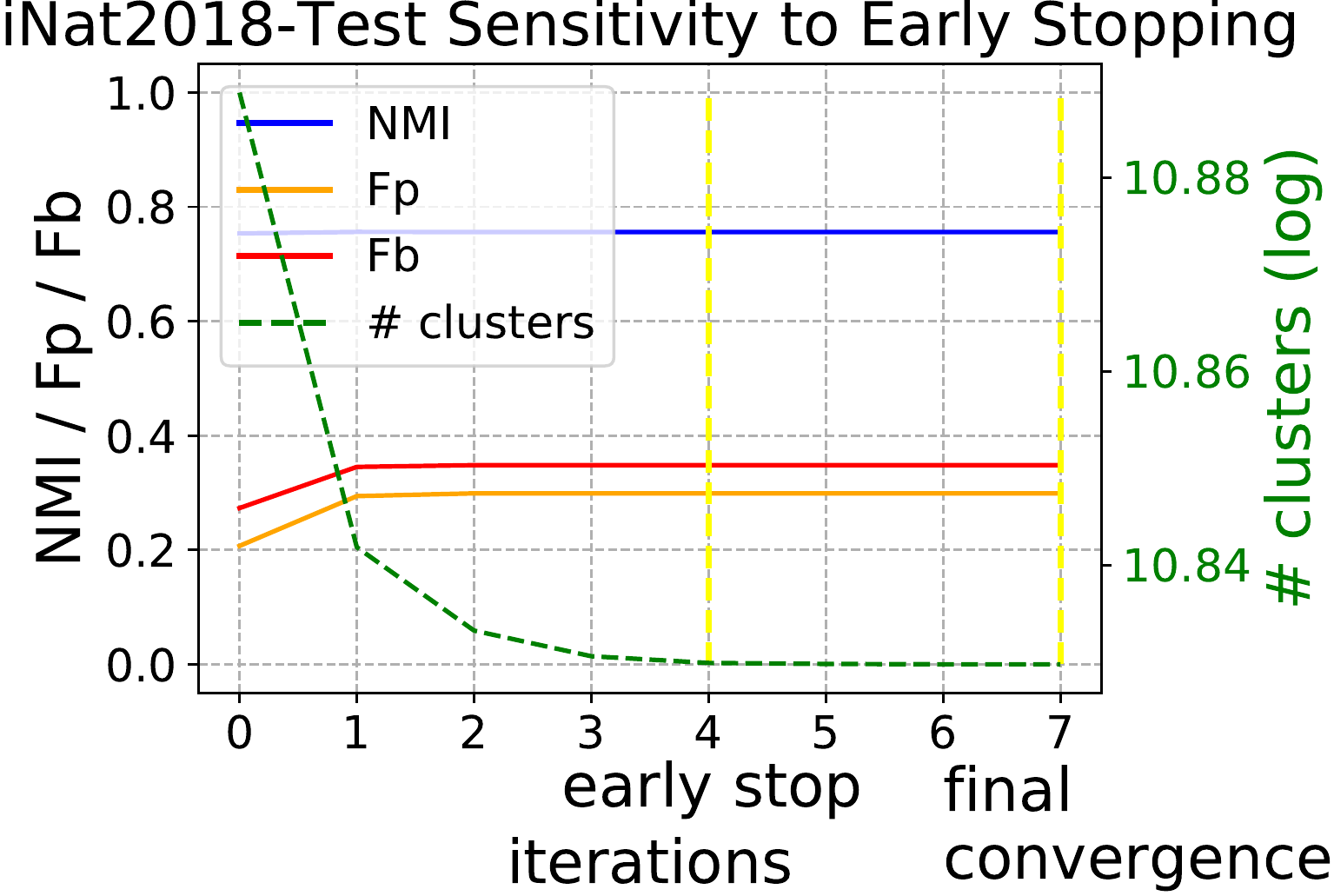}
\end{adjustbox}
                \caption{Sensitivity to early-stopping. The two dotted vertical yellow lines indicate the iterations at which the early-stopping and final convergence criteria are met. Left y-axis shows the accuracy of clustering prediction and right y-axis shows the predicted number of clusters (log-scale). The early-stopping is used to reduce further iterations after the model is close to convergence.
\label{fig:convergence_sensitivity}}
    \end{center}
    \vspace{-2em}
\end{figure}

\noindent\textbf{Sensitivity to Early-stopping}
%  \note{This is right. Earlier statement on early stopping flagged in previous section is not.}
\updateblue{The proposed agglomeration process converges when there are no more new edges added. Though this convergence is reached without an explicit termination criterion, we observe that the process can be terminated early without affecting much the final clustering accuracy.
% as shown in ~\myfigref{fig:convergence_sensitivity}. 
~\myfigref{fig:convergence_sensitivity} shows the model sensitivity to early-stopping. The two dotted vertical yellow lines indicate the iterations at which the early-stopping and final convergence criteria are met. Clustering performance  (Fp/Fb/NMI) plateaus after the iteration of early-stopping and there is no significant difference in accuracy and cluster numbers predicted compared to the final convergence. Therefore, simply for computational complexity considerations, we terminate the agglomeration early if computational cost is a concern. This choice is neither an arbitrary termination criterion nor a complexity / accuracy trade-off, rather, it is merely a computational expedient. Since there is no performance loss at early-stopping, we report performances with early-stopping in all subsequent sections.}
% There is no performance loss at the early-stopping iteration compared to the one at final convergence.
% \note{This is incorrect! The point you want to make is that there is NO tradeoff! You want to say that final performance is decided based on the inductive principle, but that performance is arrived at well before convergence occurs (see Figure 2), so you might as well reduce training time (with NO LOSS OF PERFORMANCE) by terminating performance early. Another way to say it is that even if a few edges are changing, they change in a way that does not alter performance (essentially -- just as it happens in the limit cycle path of convergence in Langevin dynamics where the solution changes, the outcome changes, but all changes lead to the same error rate - basically you are just trading errors), but it is fundamental that there be NO performance/complexity tradeoff, otherwise the whole point of not needing an arbitrary complexity threshold is out the window!}
% Ideally the model will converge when no new edges are created by the network. In practice, to ease the effect of noise, we propose a heuristic to early-stop the process. 
% \note{this simple sentence kills the pitch in the introduction. Unless we make changes, this is the precise point where the reviewer loses trust (on page 7) and decides to reject the paper. Either we change the pitch, or we show that this heuristic is just as effective as convergence to equilibrium and therefore relegate this to the implementation section, or the paper is no longer viable.} 

\updatered{Our early stopping criteria is based on the following observation. In the case where all clusters are $k$-ary trees, the number of new edges created at one level should be $\leq 1/k$ of the number of edges created in the previous level. This matches the behavior in the early hierarchies when multiple intermediate clusters are merged. In the last couple of iterations, the model adds very few number of edges for several levels before an exact convergence. Therefore, if at any level the new edges created is more than $1/k$ of the previous one, one can choose to early stop the agglomeration.}

%\note{Coudl it be that at convergence links are changing but with no effect on precision? That is, links could flip on and off in pairs for links across boundaries. If so, is there a way to do cross-validation to detemine whether the process has converged (validation error settling) as opposed to waiting for the procedure to stabilize (link labels settling)?} \yifan{the links once predicted as valid in the aggolmerative process, will stay valid from that point on-wards. When close to convergence, it's the case that there are few new edges still being added and nodes being merged, but it leads to very small changes to the final number of predicted clusters as connected components. I think it will be hard to do cross-validation as we assume there are no labels available during test-time. And different testsets, such as face video and face image (to be tested with the same ``meta-trained'' model), will have convergence at different iterations and we assume neither the validation image labels nor the distribution of ground-truth clusters on these tests are known in advance. (Unseen test-data distribution case)}}

\begin{table}[tbh]
\begin{center}
\begin{footnotesize}
\centering
    \begin{adjustbox}{max width=\linewidth}
            \begin{tabular}{c|c c c |c}
        \hline
    	 Method  & \multicolumn{3}{c}{Hannah} & Runtime\\
    	\hline
    	& Fp & Fb & NMI & sec \\
    % 	\hline
    % 	\hline
    % 	GCN-V+E \cite{GCN-VE} & 0.918 & 0.491 &	0.640 & 256.2\\
    % 	\hline
    % 	\hline
    %     \NAME w.o. GAT  & 0.900 &	0.498	&0.641  & 42.7 \\
    %     \NAME w. GAT  & 0.954 &	0.488	& 0.645 & 44.9 \\
    %     \HNAME  & 0.933 & 0.695 & 0.796 & 36.9 \\    
    
        \hline
    	\hline
    	GCN-V+E \cite{GCN-VE} & 0.062 & 0.224 &	0.640 & 256.2\\
        \NAME & 0.065 &	0.234	& 0.644 & 44.9 \\
        % \HNAME & \textbf{0.720}	& \textbf{0.682}	& \textbf{0.798} & \textbf{36.9} \\    
        \HNAME & \textbf{0.720}	& \textbf{0.700}	& \textbf{0.810} & \textbf{36.9} \\ 
    	\hline
    \end{tabular}
    \end{adjustbox}
    \caption{\label{tbl:ablation_gat} Ablation experiments:1) value of joint prediction compared to inference with two-separate models 2) value of hierarchy.}
\end{footnotesize}
\vspace{-1.5em}
\end{center}
\end{table}
\noindent\textbf{Value of Joint Inference} We examine the effect of joint inference in our single level \NAME model compared to a prior GNN \cite{GCN-VE} that uses two-separate models in \mytabref{tbl:ablation_gat}. The joint model outperforms the baseline, with a 5\% boost in F-scores, while reducing the runtime by five-fold.

\noindent\textbf{Value of Hierarchy Design} We examine the effect of the hierarchical design in \HNAME in \mytabref{tbl:ablation_gat}. 
% The motivation of LANDER is to find a joint model that is versatile in both link prediction and density estimation while remaining computationally efficient. 
Comparing row two and three, combining \NAME with our hierarchical approach resulting in \HNAME brings significant gains in F-scores from 0.234 to 0.700 and increase in NMI from 0.644 to 0.810 via \updateblue{modeling the data granularity using a disjoint meta-training set with a learnt convergence.}

\begin{table}[tbh]
\begin{center}
\begin{footnotesize}
\centering
    \begin{adjustbox}{max width=\linewidth}
            \begin{tabular}{c |c c c |c c c}
    	\hline
    	 Method  & \multicolumn{3}{c}{IMDB-Test-SameDist} & \multicolumn{3}{c}{iNat2018-Test} \\
    	\hline
    	& Fp & Fb & NMI & Fp & Fb & NMI \\
    	\hline
    	\hline

    	DBSCAN \cite{DB-SCAN}   & 0.064 &	0.092	& 0.822 & 0.100 &	0.116&	0.753 \\
    	
    	ARO \cite{ARO} &0.012&	0.079&	0.821 & 0.007&	0.062&	0.747 \\
    	\hline
    	HAC \cite{HAC} &0.598  & 0.591 & 0.904 & 0.117 &	0.245 & 0.732	 \\
    	
    	H-DBSCAN \cite{H-DBSCAN} & 0.423&	0.628	&0.895 &0.178 &	0.241&	0.754 \\
    	
    	Graclus \cite{Graclus} & 0.014	&0.099&	0.829 & 0.003	&0.050	&0.735 \\
    	
    	FINCH \cite{FINCH} & 0.001	& 0.001 & 0.155 & 0.014 &	0.014 & 0.283 \\
        \hline
    	
    	LGCN \cite{L-GCN}  & 0.695 & 0.779  & 0.940 & 0.069 & 0.125  & 0.755 \\
    	
    	GCN-V \cite{GCN-VE} & 0.722&	0.753&	0.936& 0.300 &	\textbf{0.360}&	0.719 \\
    	
    	GCN-V+E \cite{GCN-VE} & 0.345	&0.567	&0.864 & 0.273 &	0.353&	0.719 \\

    	\hline
    	\hline
    % 	\HNAME & \textbf{0.770} &\textbf{0.784} & \textbf{0.944} & \textbf{0.330}	& 0.350	& \textbf{0.774} \\
    \HNAME & \textbf{0.793} &\textbf{0.795} & \textbf{0.947} & \textbf{0.330}	& 0.350	& \textbf{0.774} \\

    	\hline
    \end{tabular}
    \end{adjustbox}
    \caption{\label{tbl:same_domain_train_test} \updateblue{Same train-test distribution clustering performance. First six rows show the unsupervised baselines (latter four are hierarchical based) and the last four rows show the supervised GNN based methods (including ours). \HNAME outperforms both prior SOTA unsupervised and supervised GNN methods, with an average improvement of 37\% and 5\% in F-scores, respectively.}}
    %  \note{useless caption. It just duplicates the header. What am I supposed to surmise from this experiment? What is the message it conveys?}
\end{footnotesize}
\vspace{-2em}
\end{center}
\end{table}

\subsection{Clustering Performance\label{sec:same_train_test}}
Here, we compare \HNAME with state-of-art unsupervised and supervised methods under the setting where the cluster size distributions of train and test data are similar. For face, we sample a subset of IMDB to match the training distribution of Deepglint, and name this sub-sampled testset as IMDB-Test-SameDist. For species, we use iNat2018-Train and iNat2018-Test for training and testing since they follow the same cluster size distribution. \updateblue{~\mytabref{tbl:same_domain_train_test} shows the results. \HNAME consistently outperforms prior SOTA unsupervised and supervised GNN baselines, with an average boost of 37\% and 5\% in F-scores, respectively.
% \note{cite the numbers, for instance "uniformy, with an average improvement of X\%}. 
% This demonstrates the high clustering prediction accuracy of the proposed hierarchical GNN meta-clustering method. 
Supervised baselines performs better than unsupervised ones in this setting. We hypothesize that this is due to the domain specialization in dealing with complex cluster structure through GNN training on label-annotated datasets.}

\begin{table}[tbh]
\begin{center}
\begin{footnotesize}
\centering
    \begin{adjustbox}{max width=\linewidth}
            \begin{tabular}{c |c c c |c c c| c c c }
        \hline
    	 Method  & \multicolumn{3}{c}{Hannah} & \multicolumn{3}{c}{IMDB} & \multicolumn{3}{c}{iNat2018-Test} \\
    	\hline
    	& Fp & Fb & NMI & Fp & Fb & NMI & Fp & Fb & NMI \\
    	\hline
    	\hline
    	
    	DBSCAN \cite{DB-SCAN} &	0.041	& 0.128 & 0.546  & 0.057 &	0.118 & 0.851   &	0.100 &	0.116 & 0.753 \\
    	
    	ARO \cite{ARO} & 0.001	 & 0.018  & 0.483 &	0.012 &0.103& 0.849 & 0.007&	0.062&	0.747 \\
    	\hline
    	
    	HAC \cite{HAC} & 0.197 &	0.475 & 0.521 &	0.592 & 	0.624 & 0.923 & 0.117	&0.245 &0.732\\
    	
    	H-DBSCAN \cite{H-DBSCAN} &	0.112	& 0.296  & 0.526 &	0.395 &	0.641 &  0.912& 0.178 &	0.241&	0.754 \\
    	
    	Graclus \cite{Graclus} & 	0.001	 &0.004 &  0.452 & 0.018 &	0.131 & 0.857 & 0.003	&0.050	&0.735 \\
    	
    	FINCH \cite{FINCH} 	& 0.265 &	0.258 & 0.338 & 0.001 & 0.001 & 0.089 & 0.014 & 0.014 & 0.283\\
        \hline
    	
    	LGCN \cite{L-GCN}  & 0.002 &	0.098& 0.455 & 0.665 &	0.771&  0.946& 0.030&	0.076&  0.747\\
    	
    	GCN-V \cite{GCN-VE} &	0.056 &	0.218 & 0.637 & 	0.634 & 0.768 & 0.948  & 0.269	& \textbf{0.352} & 0.719	 \\
    	
    	GCN-V+E \cite{GCN-VE} & 	0.062 & 	0.224 & 0.640 &	0.589&	0.732 & 0.940 & 0.252 & 0.338 & 0.719\\

    	\hline
    	\hline
    % 	\HNAME & \textbf{0.720}	& \textbf{0.682}	& \textbf{0.798} & \textbf{0.717} & \textbf{0.786} &	\textbf{0.950}	 &	\textbf{0.294} & \textbf{0.352} & \textbf{0.764} \\
    \HNAME & \textbf{0.720}	& \textbf{0.700}	& \textbf{0.810} & \textbf{0.765} & \textbf{0.796} &	\textbf{0.953}	 &	\textbf{0.294} & \textbf{0.352} & \textbf{0.764} \\
    	\hline

    \end{tabular}
    \end{adjustbox}
    \caption{\label{tbl:diff_domain_train_test} Clustering with unseen test data distribution. iNat2018-Test results of the supervised methods are from models trained on iNat2018-Train-DifferentDist. \updateblue{\HNAME outperforms SOTA GNN supervised and unsupervised methods, with an average F-score boost of 54\% and 51\% respectively. On Hannah, where the test-distribution is very different from that in meta-training, we improve the F-score from 0.224 to 0.700 over prior GNN methods.}} 
    % \yifan{Update iNat to use the Focal loss results} }
    % \david{Perhaps it would be better to remove one digit of precision from these results, consistent with Table 4, such that the table size could be expanded a bit (now seems overly compressed).} \note{+1} this is unreadable. Also the caption needs to be informative. It needs to explain what the numbers show and what the reader is supposed to take away from this table. What should I conclude from these numbers?
\end{footnotesize}
\vspace{-2em}
\end{center}
\end{table}

\subsection {Clustering with Unseen Test Data Distribution}
We also report clustering performance under the setting where test-time distribution is unknown and different from that of meta-training. Namely, parameters (such as $\tau$ and $k$-NN $k$ in GCN-V/E and max cluster size in L-GCN) cannot be adjusted in advance using test-time information. For face clustering, we train with TrillionPairs-Train and test on Hannah and IMDB. For species, we sample a subset of the iNat2018-Train to attain a drastically different train-time cluster size distribution as iNat2018-Test, which is named as iNat2018-Train-DifferentDist. ~\mytabref{tbl:diff_domain_train_test} illustrates the results. \HNAME outperforms prior SOTA supervised baselines, with an average 54\% F-score boost. Over Hannah, \HNAME achieves a significant F-score improvement from 0.224 to 0.700 and a NMI boost from 0.640 to 0.810, demonstrating its strong generalization capability to unseen test distributions. Some unsupervised baselines such as H-DBSCAN and HAC outperform the supervised baselines over Hannah, showing better generalization capability. Despite being a supervised method, \HNAME outperforms all unsupervised baselines across different test sets, with an average 51\% boost in F-scores, owing to the strong expressive power of our unified GNN \NAME model.

\begin{table}[tbh]
\begin{center}
\begin{footnotesize}
\centering
    \begin{adjustbox}{max width=\linewidth}
            \begin{tabular}{c |c c |c c }
    \hline
    	 Method  & \multicolumn{2}{c}{IJBC 1:1 FNMR@FMR} & \multicolumn{2}{c}{IJBC 1:N FNIR@FPIR} \\
    	\hline
	& 1e-3 &	1e-4 & 	1e-1 &	1e-2 \\
	\hline
	\hline
    Graclus \cite{Graclus}	& 0.290	& 0.461 & 0.467	& 0.620 \\
    FINCH \cite{FINCH}	& 0.133	& 0.230	& 0.240 &	0.375 \\
    H-DBSCAN \cite{FINCH}& 0.111	& 0.200 & 	0.196	& 0.312 \\
    \hline
    GCN-V \cite{GCN-VE}& 0.107	&0.181	&0.181	&0.270 \\
    GCN-V+E	\cite{GCN-VE} & 0.110&	0.187	&0.191	&0.291 \\
    \HNAME &\textbf{0.091}&	\textbf{0.159}&	\textbf{0.162}	& \textbf{0.250} \\
    \hline
    Fully-supervised &0.072	&0.136	&0.136	&0.235 \\
    	\hline
    	\hline
    \end{tabular}
    \end{adjustbox}
    \caption{\label{tbl:downstream_face} \updateblue{Face recognition on IJBC\cite{IJBC}. \HNAME outperforms all baselines and improves over the best result from prior-arts with a 14\% error reduction. Models trained with pseudo labels generated by \HNAME brings the performance (0.159) closer to the lower bound \tong{Maybe we need citation for this supervised benchmark} with fully supervised training (0.136).}}
        % \david{but isn't this a lower bound, meaning the lowest possible value of the metric being used?} \note{+1: good performance = small number. You can just say "the paragon" to avoid confusion on the direction.} 
    % \david{lower bound?} \yifan{lower bound is usually defined in setting where a mixture of labeled and unlabeled data are used for training recogntion, here we use purely unlabeled data, so hard to define a lower bound in this setting.}
\end{footnotesize}
\vspace{-2em}
\end{center}
\end{table}

\subsection{\updateblue{Representation Learning with Pseudo-Labels}}
% We further demonstrate the advantage of \HNAME in generalization capability through the task of face recognition with pseudo label training. 
We follow a setting similar to that of \cite{CDP, roychowdhury2020improving, GCN-DS} for face recognition with pseudo label training. Starting with an initial representation learned through some labeled datasets,
%  After which, a large-scale unlabeled dataset is utilized to better learn a representation through pseudo-label generation.
we utilize the clustering methods to generate pseudo labels for unlabeled datasets and train with these pseudo labels to better learn a representation.\footnote{The pseudo-labels generated by our clustering, or any other deterministic or stochastic processing of the training set, span the same Sigma Algebra as the training set, so they cannot be thought of as ``ground truth'' or ``additional information'' when training a classifier with pseudo-supervision. However, the pseudo-labels capture the inductive bias of the training process and therefore serve as a regularizer that, while not adding information, nevertheless improves generalization, as shown empirically.} 
% \note{It could be OK to move the footnote in the introduction here.}
The face recognition experiment involves the following steps: 1) Start with an initial face recognition model learnt on TrillionPairs. 2) Train a clustering model on the TrillionPairs or use an unsupervised clustering method with the initial face representations. 3) Generate pseudo label on IMDB (overlapping identities with TrillionPairs removed). 4) Train face recognition models on IMDB via the pseudo labels. 5) Evaluate the learned face representation on the open-set IJBC benchmark. \updateblue{
~\mytabref{tbl:downstream_face} shows the results. We also report the lower bound from fully supervised training on IMDB with human labeled data. \HNAME achieves a 14\% error reduction compared to the best baseline. Interestingly, pseudo label training with \HNAME brings the performance to 0.159 (verification FNMR@FPIR1e-4), closer to the lower bound of fully supervised training at 0.136 than any of the baselines.}

\subsection{Runtime Analysis}

\begin{table}[tbh]
\begin{center}
\begin{footnotesize}
\centering
     \begin{adjustbox}{max width=\linewidth}
            \begin{tabular}{c | c | c | c }
        \hline
    	 \diagbox{Method}{Dataset} & Hannah & IMDB & iNat2018-Test\\
        \hline
        DBSCAN \cite{DB-SCAN}  & 480.2 & 10,358.0 & 592.6 \\
        \hline
        ARO  \cite{ARO}   & 184.4 & 1,349.3 & 223.9 \\
        \hline
        \hline
        HAC \cite{HAC}    & 446.8   & 183,311.8 & 6,730.5  \\
        \hline
        H-DBSCAN \cite{H-DBSCAN} & 9,865.3  & 390,360.0 & 121,821.0 \\
        \hline
        Graclus \cite{Graclus} & 38.3 & \textbf{176.6} & 47.4 \\
        \hline
        FINCH \cite{FINCH} & 74.7 & 300.4 &	\textbf{46.2}\\
        \hline
        \hline
        LGCN \cite{L-GCN}& 3,342.1 & 33,211.1 & 3,057.4\\
        \hline
        GCN-V  \cite{GCN-VE}& 41.7 & 204.8 & 53.4 \\
        \hline
        GCN-V+E \cite{GCN-VE}  & 256.2 & 3,283.3 & 197.5  \\
        \hline
        \hline
        \HNAME & \textbf{36.9} & 511.0 & 67.4  \\
    	\hline
    % 	\HNAME@Level & 36.9@5 & \updategreen{511.0@3} & 67.4@2  \\
    \end{tabular}
    
    %     \begin{tabular}{c | c || c | c  |c ||c |c | c | c || c | c}
    %     \hline
    % 	 \diagbox{Dataset}{Method} & \HNAME & LGCN & GCN-V & GCN-V+E & HAC & H-DBSCAN & Graclus& FINCH & DBSCAN & ARO \\
    %     \hline
    %     Hannah & 36 & 3,342& 41&256  & 446 & 9865& 38 & 74& 480&  184\\
    %     \hline
    %     IMD & 511& 33,211& 204&3,283 & 183,311 & 390360 & 176& 300& 10,358& 1,349 \\
    %     \hline 
    %     iNat2018-Test & 67& 3,057& 53& 197&  6,730& 121821 & 47& 46 & 592 & 223  \\
    % 	\hline
    % % 	\HNAME@Level & 36.9@5 & \updategreen{511.0@3} & 67.4@2  \\
    % \end{tabular}
    \end{adjustbox}
    \caption{\label{tbl:all_runtime} Runtime comparison on all benchmarks in \mytabref{tbl:diff_domain_train_test} (secs).}
\end{footnotesize}
\vspace{-2em}
\end{center}
\end{table}

% \begin{figure}[t]
%     \centering
%     %\includegraphics[width=0.9 \linewidth]{figures/hannah_runtime_across_k.eps}
%     \includegraphics[width=0.9 \linewidth]{figures/hilander-iccv-runtime-nmi.png}
%     \caption{\updateblue{Cost-complexity comparison with state-of-the-art GNN clustering \cite{GCN-VE} on Hannah \cite{Hannah}. Illustrating improvement in performance with simultaneous reduction in cost. TO BE DECIDED if we put in this or remove to avoid confusion on k-tuning.}}
%     % \note{This is not run-time  comparison. It is the cost/performance curve of two algorithms on a specific datasaet. If this is the closest kin to compare with, you can move this to the first page with the title "cost-complexity comparison with state-of-the-art GNN clustering, illustrating improvement in performance with simultaneous reduction in cost.}
%     \label{fig:hannah_runtime}
% \end{figure}

% \begin{figure}[tbh]
%     \begin{center}
%     \centering
% \begin{adjustbox}{max width=\linewidth}
%         \includegraphics[width = 1.0\textwidth]{./figures/hannah_dive_deeper.pdf}
% \end{adjustbox}
%                 \caption{\HNAME analysis on Hannah with multi-class visualization: We show three different classes that converge at different levels of the proposed hierarchical graph clustering method.\label{fig:hannah_deeper}}
%     \end{center}
% \end{figure}

% We measure the runtime with 8-core Intel(R) Xeon(R) E5-2686 v4 CPU and Tesla V100 GPU. Our proposed models use PyTorch\cite{pytorch} v1.5, DGL\cite{dgl} v0.5 with CUDA v10.1. $k$-NN building leverage faiss\cite{faiss}
We compare the runtime (seconds) of \HNAME with all baselines (\mytabref{tbl:all_runtime}). The hardware and software specifications are included in the appendix. The complexity numbers above are from \HNAME with early-stopping. Our method is faster than most baselines and is comparable with GCN-V\cite{GCN-VE}, FINCH\cite{FINCH} and Graclus\cite{Graclus}. The multiple hierachical levels introduced do not bring in additional overhead since \HNAME runs faster level by level, with fewer number of nodes remaining after each level.

% We further demonstrate the advantage of \HNAME in cost-complexity by comparing to GCN-V\cite{GCN-VE} whose parameter $k$ is illegally tuned using unseen test data distribution. Even if the test-time information is leaked to the baseline method, we outperform its accuracy and at the same time being 20$\times$ faster (\myfigref{fig:hannah_runtime})

% Our framework is slower when $k$ in their model is small. The gap is gradually filled with the increase of $k$ in GCN-VE. We will have both NMI advantage and runtime efficiency after $k > 160$. 

% However, it brings significant runtime benefit compared with GCN-E. Specifically, GCN-E runs inference on a 1-hop neighbor subgraph for each node sequentially. Though it can be run in parallel, it cannot aggregate information beyond the sampled sub-graph and thus requires more GCN layers than our design to obtain the same amount of information. Furthermore, our model can run in full graph mode and batch sampling mode, better utilizing the parallelism of computing devices.

\section{Discussion \label{sec:discussion}}

% \note{To be re-done. This section should not be a summary of the paper (that is what the abstract is for), nor should be a repetition. It should be a discussion of the lessons learned, an honest assessment of the limitations, and a highlight of failure cases. CUT THE FOLLOWING -- This paper proposes a supervised hierarchical graph neural network visual clustering method to tackle the issue of limited generalization capability in existing supervised GNN based clustering methods. In addition, a novel unified joint graph linkage prediction and density estimation model has been proposed for efficient and accurate clustering performance. Experiments over multiple tasks of visual clustering and downstream representation learning validates the effectiveness of the method in generalizing to unseen data distributions. It is also demonstrated that the end-end method, \HNAME, achieves state-of-the-art performance on multiple large-scale visual clustering benchmarks.}

% \note{list here all the conclusions you draw from the experiments -- which you could not have discussed in the introduction, focusing on limitations and failure cases.} 
The proposed clustering method aims at providing a rich representation of unlabeled data using induction from an annotated training set. GNNs represent a natural tool, for they allow training from a disjoint dataset a model that outputs a graph structure. Since the clustering problem is intrinsically ill-posed, for there is no unique ``true'' cluster, we aim to provide a rich hierarchical representation that gives the user more control -- in the spirit of agglomerative hierarchical clustering. To tackle the computational challenge in replicating the basic graph operations across levels of the hierarchy, we have proposed enhancements of current GNN-based methods that improve efficiency. \updateblue{Though the complexity of our method is $\mathbb{O}(kN)$, the same as the vanilla flat-version of GNN clustering, the full-graph inference is a natural parallelization and significantly reduces the runtime compared to prior GNNs with sub-graph inference.}
% by  $\mathbb{O}(N)$
% the complexity of our method is \note{describe}, which is an increase/decrease of X\% compared to the vanilla flat-version of meta-clustering with a GNN.

\HNAME is subject to the usual failure modes of all inductive methods, when the distribution of test data is extremely different from that in training. \updateblue{In addition, the current node feature aggregation takes the form of averaging while there might be more sophisticated methods such as learnable attention for more informative aggregation.}
% , \note{describe specific limitations of the method}.

% Our method entails a number of design parameters. Some are unavoidable and even beneficial for unsupervised clustering, again because there is no independent validation mechanism and therefore it is beneficial -- if not necessary -- to provide the user with tuning parameters to calibrate the model for a particular use case. \updateblue{Other parameters include base cluster and aggregation function choices. We have conducted sensitivity experiments to quantify the effect of these choices on the output set of clusters, and have revealed that \HNAME is robust to these design choices.} 
Even so, our goal is to reduce the number of arbitrary choices as much as possible and defer to the data the most critical design decisions. One is the choice of clustering criterion. This is inherited by the training set, through the simple classification loss. So is the level of granularity of the partition of the data.
%  used during training
% So is the model complexity, or correspondingly the level of granularity of the partition of the data, or data complexity.
Though we use early stopping, we do so only after verifying that the method, when iterated to convergence, settles on a solution that is not substantially different from that obtained in earlier iteration.  Therefore early stopping is not chosen as a design parameter or an inductive bias, but merely as a way to reduce computation. %as a computational expedient to keep computational cost at bay. 
% \note{the last statement to be verified}

% \yifan{Where do we discuss the diff to GCN-V/E paper? currently it is in the related work and contribution section on supervised GNN clustering.}

% \tianjun{teaser could be a demonstration of different domain train/test}

% \yifan{teaser put the hierarchical GNN inference design? we can delay the teaser decision later, giving higher priority to the paper length.}

\clearpage
\appendix
\section*{Appendix}
\section{Hi-LANDER Clustering Visualization}
We visualize in \myfigref{fig:hannah_deeper} the hierarchical clustering process of the proposed method \HNAME.
% We demonstrate in \myfigref{fig:hannah_deeper} the effectiveness of the proposed method \HNAME in dealing with test-time classes having a heavily long-tail distribution in number of instances per class. 
We show three ground-truth clusters that differ in cluster sizes and embed their features into a 2D plane with t-SNE, and then visualize the points (as shown on the left column). The blue squares are the input nodes at each level of the hierarchy. The colored dots are peak nodes that are grouped from the intermediate clusters (connected-components), and the colors represent the three different ground-truth classes. Note that the peaks at each level then become ordinary input nodes at the next level.

We see that the nodes in the red cluster are grouped efficiently with only one peak node left in level 1, while there are many small clusters for the yellow and green class nodes. In the next hierarchy, as shown in the second row, the distance between each pair of the peak nodes is larger, and the number of peaks reduced rapidly. The red cluster stays unchanged since our base clustering model \NAME stops adding edges, while the green and yellow clusters are further grouped. The last row shows the final level where all three classes converge, and only nodes belonging to the yellow cluster are further grouped.

%The top-left figure shows the first hierarchy of the three classes. Given $k = 10$, we see that the nodes in the red cluster are grouped efficiently with only one peak node left, while there are many small clusters for the yellow and green ones.

%The top-right figure shows the next hierarchy, and we can see that the distance between each pair of the peak nodes are larger, and the number of peaks reduced rapidly. The red cluster keeps unchanged since our link approximator will stop adding edges to that cluster, while the green and yellow clusters are further grouped.

%The two figures in the second row show the next two hierarchies, and only nodes belong to the yellow cluster are further grouped.

Besides, on the right column of \myfigref{fig:hannah_deeper}, we demonstrate the actual face images corresponding to the peak nodes at each level of the hierarchical clustering process for all three classes. Compared to level 2 peaks, the images corresponding to level 1 peaks are more ``repetitive.'' If we run a prior GNN based clustering model that only produces a single partition, each ``repetitive'' level 1 peak will lead to a separate cluster, and this results in low clustering completeness. In level 2, the large number of small clusters corresponding to the yellow class are grouped into 4 larger clusters. As shown in the second row of the right column in \myfigref{fig:hannah_deeper}, the images correspond to the peak nodes of these 4 clusters (with the yellow boundary) become less visually similar, while one can tell that they still represent the same person. Note that the three classes converge at different levels. Nodes of the red class already converge at the first level, the green class nodes converge at level2 while the yellow class requires all three levels to reach convergence. This illustrates the variance of real-world test data where the instance per class can be very different from class to class and it demonstrates \HNAME's capability in dealing with such large variance.

\begin{figure}[tbh]
    \begin{center}
    \centering
\begin{adjustbox}{max width=\linewidth}
        \includegraphics[width = 0.5\textwidth]{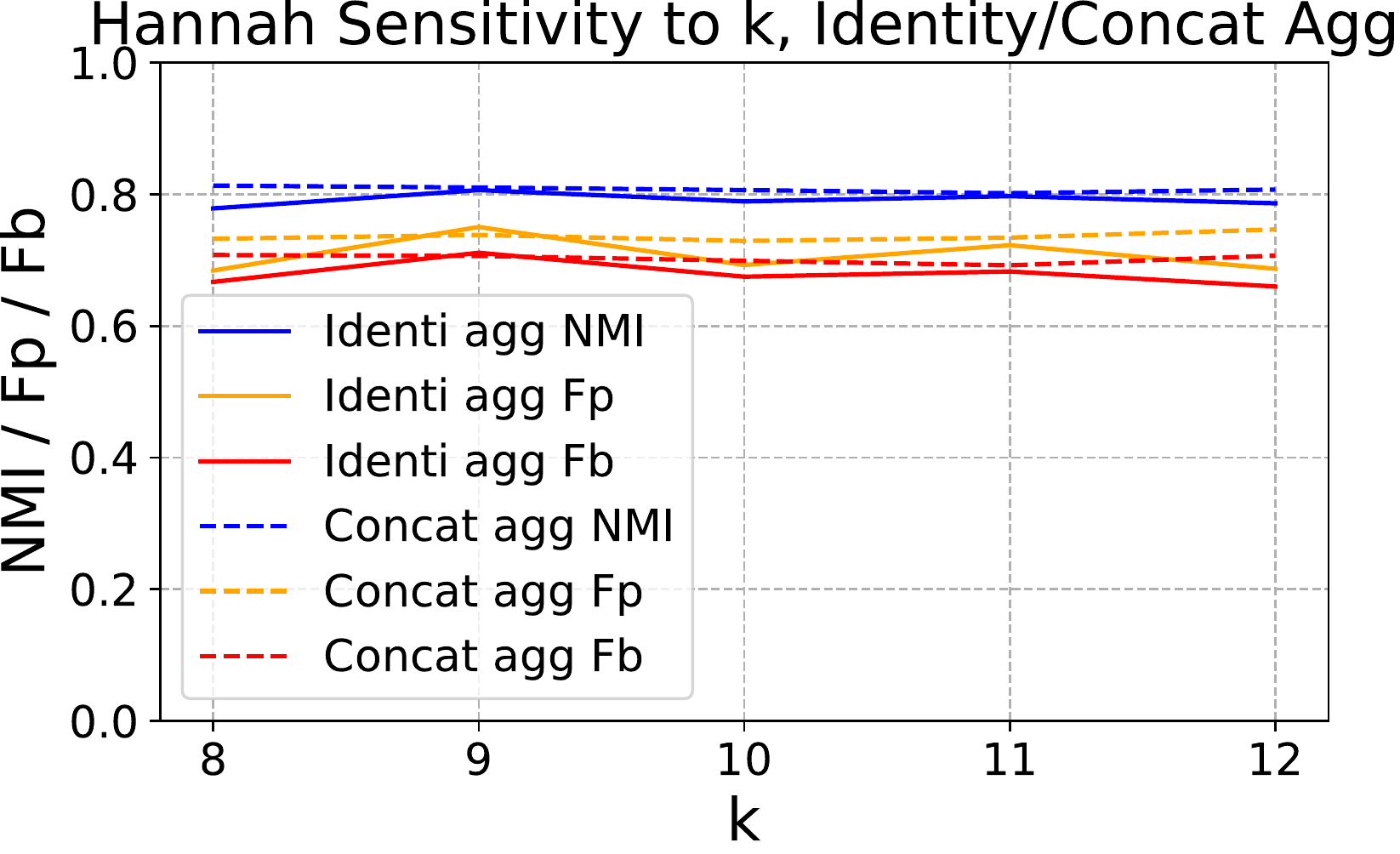}
        \includegraphics[width = 0.5\textwidth]{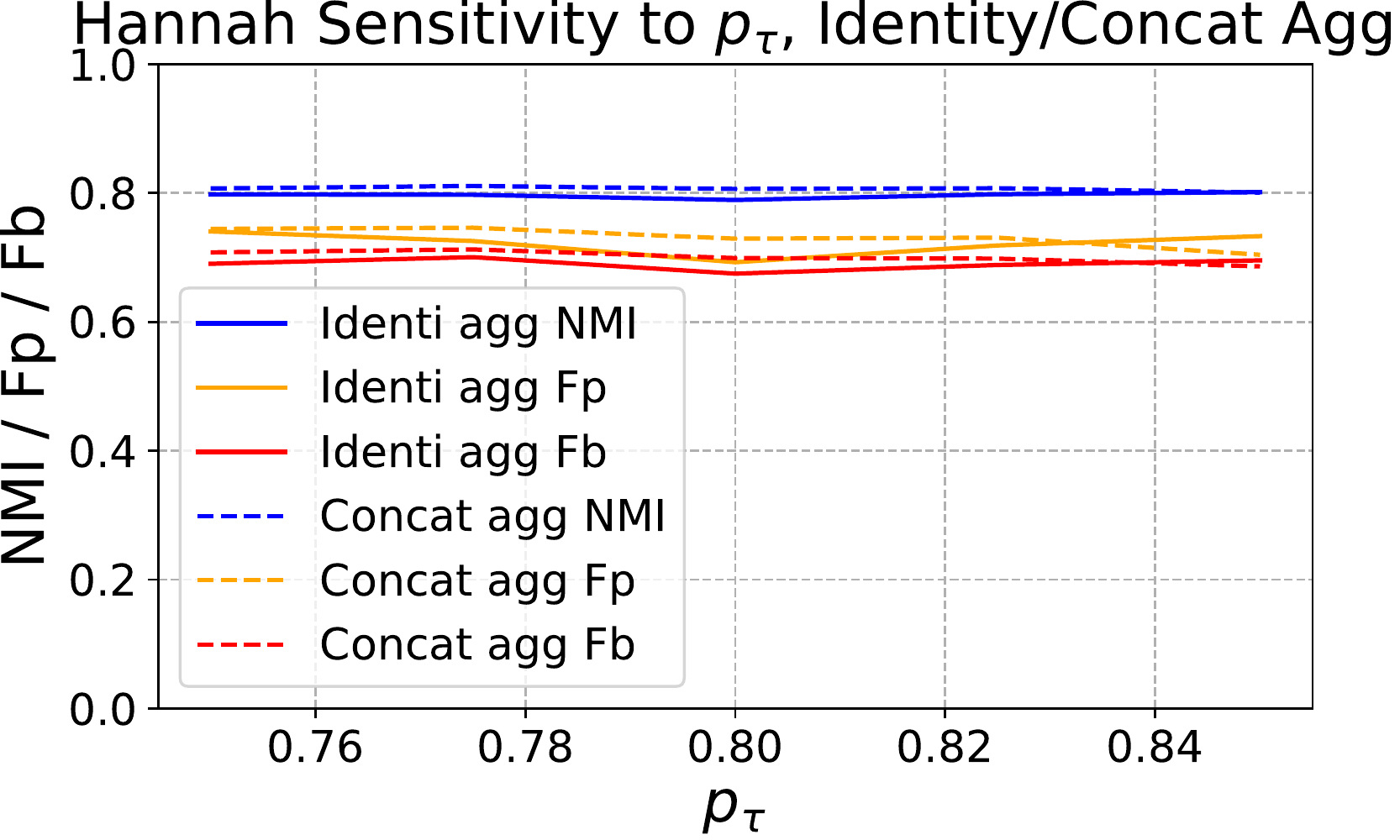}
\end{adjustbox}
\begin{adjustbox}{max width=\linewidth}
        \includegraphics[width = 0.5\textwidth]{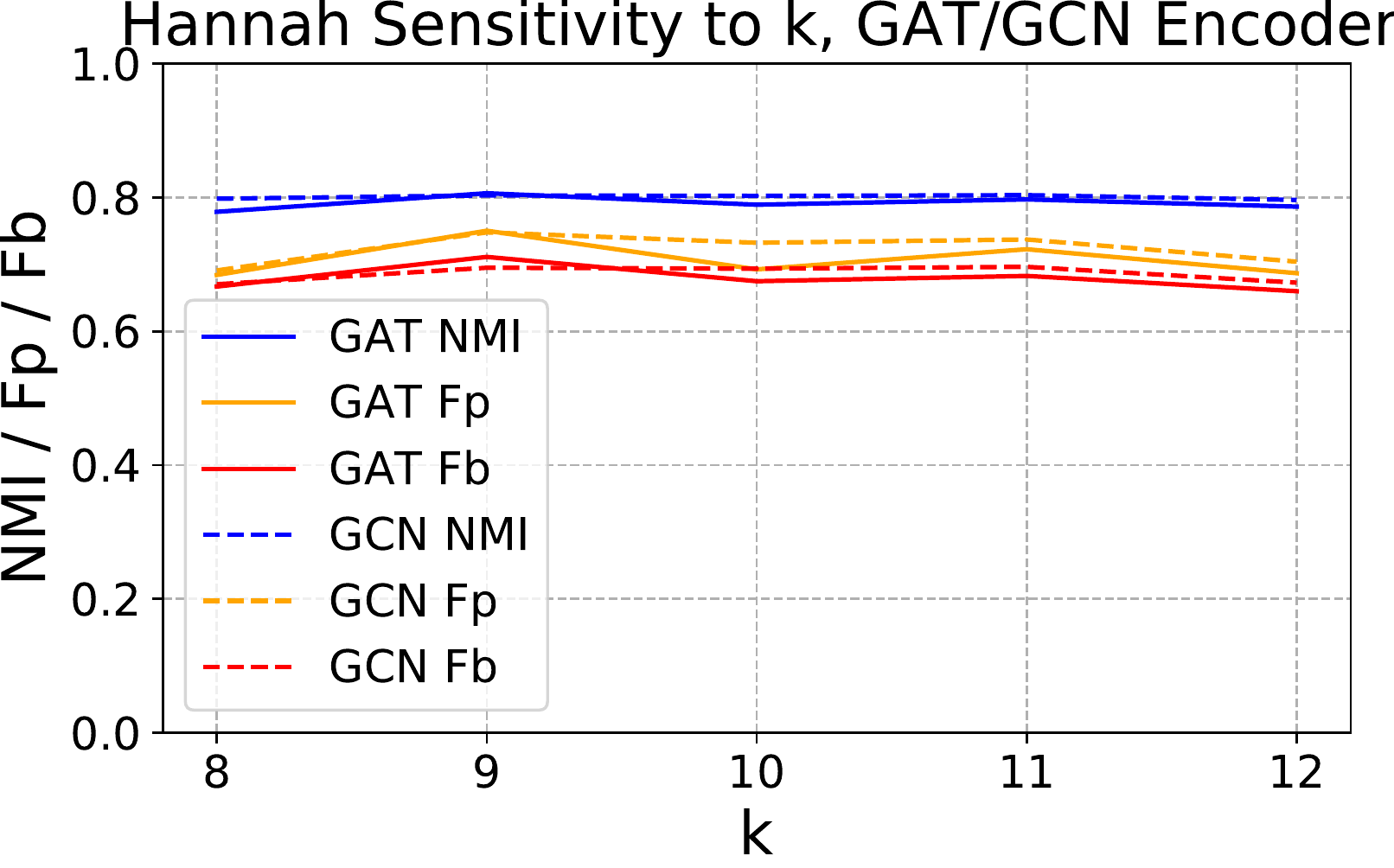}
        \includegraphics[width = 0.5\textwidth]{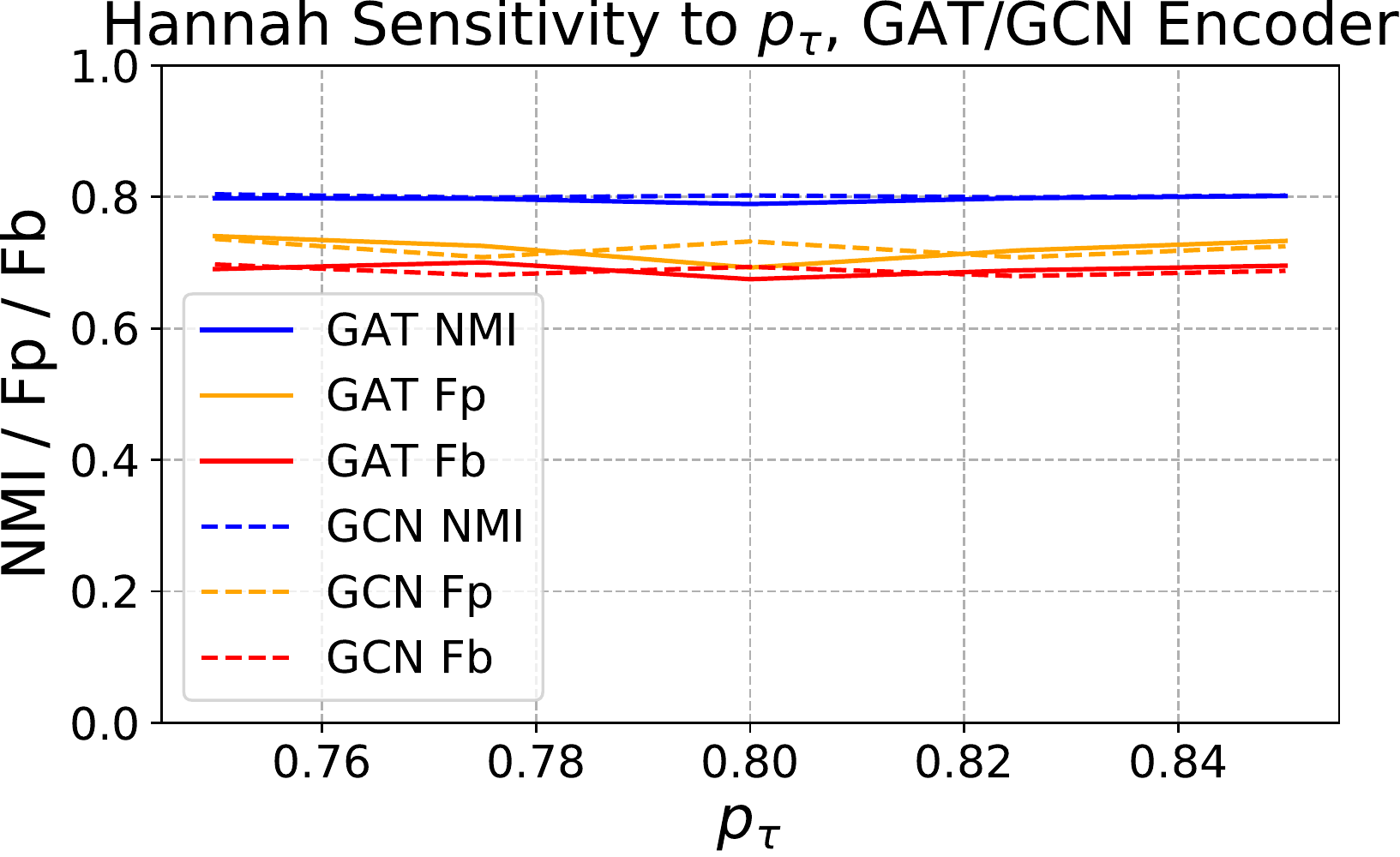}
\end{adjustbox}
                \caption{Sensitivity to hyper-parameters (chosen on the validation set, a part of the meta-training set) on the Hannah (unseen cluster-size distribution to that in meta-train) face clustering benchmark. The top two plots show sensitivity of \HNAME to the hyper-parameters of $k$, $p_\tau$ and the feature aggregation mechanism, where solid lines show the results of identity feature aggregation and dashed lines show the results from concatenation of identity and average feature. The bottom two plots show sensitivity to different types of encoders, a GAT layer (solid lines) compared to a vanilla GCN layer (dotted lines). For $k$ sensitivity tests, we vary it around the optimal value of 10 from 8 to 12. For $p_\tau$ sensitivity tests, we vary it around the optimal value of 0.8, from 0.75 to 0.85 with interval of 0.025. All three clustering metrics of NMI (blue), Fp (yellow) and Fb (red) are shown. Best viewed in color.
\label{fig:k_sensitivity}}
    \end{center}
\end{figure}

% More levels may conduct further grouping and help to improve the completeness.
% Lastly, we also show the images of the clusters corresponding to two other identity classes in \myfigref{fig:hannah_other_classes} at their respective convergence level. Due to the large variance in ground truth cluster sizes, different classes might have different sensitivity to $k$. For example, in \myfigref{fig:hannah_other_classes}, the red class converges simply at the first level of the hierarchy while the identity class referred by the green color needs two hierarchies to converge. 

\begin{figure*}[tbh]
    \begin{center}
    \centering
\begin{adjustbox}{max width=\linewidth}
        \includegraphics[width = 0.95\linewidth]{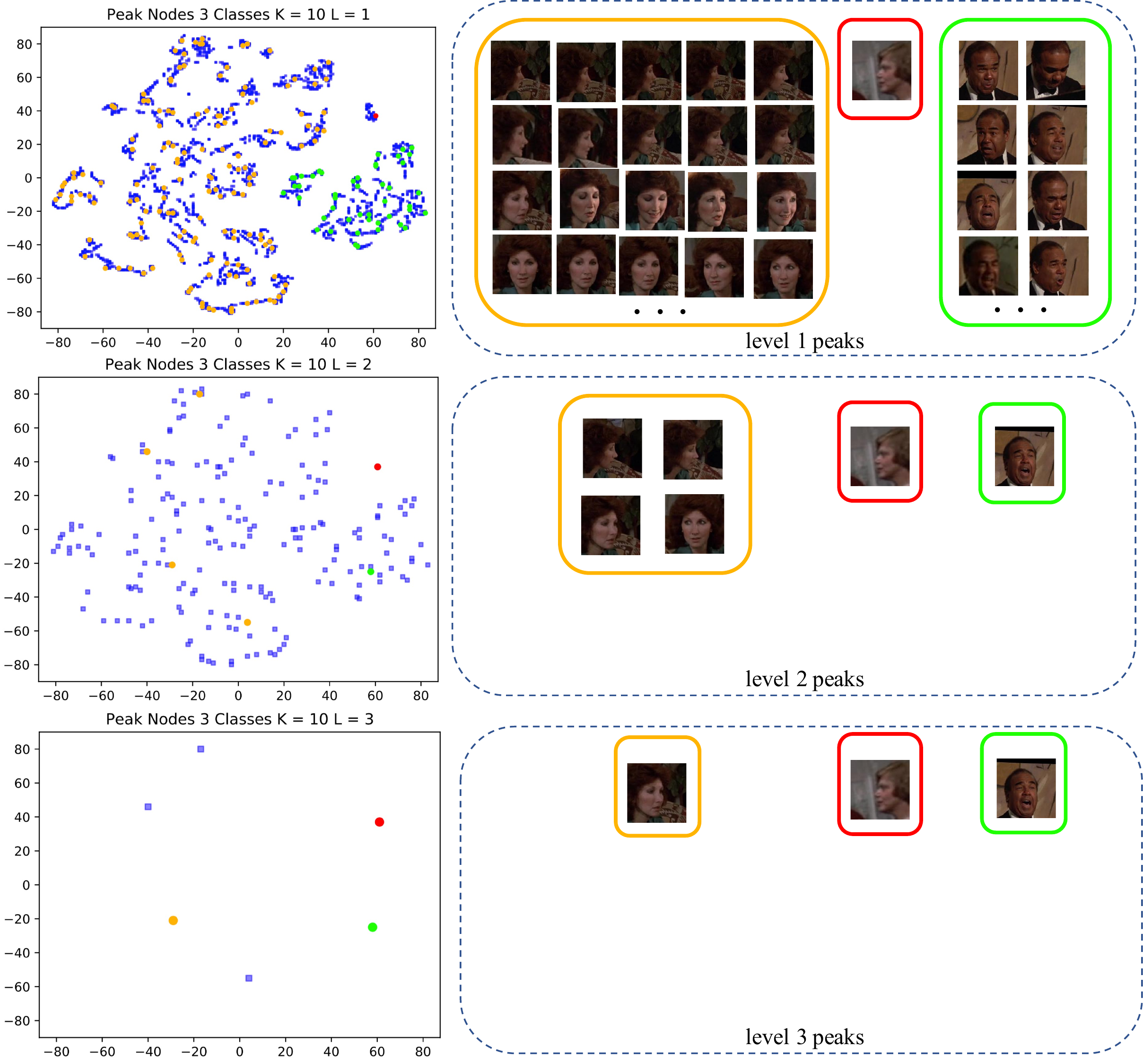}
\end{adjustbox}
                \caption{\HNAME clustering process visualization on Hannah with multiple image classes. The yellow, red and green color represent three different classes that vary in cluster size. The left column shows the t-SNE \cite{vanDerMaaten2008} embedded nodes and peaks from level 1 to level 3 of \HNAME's hierarchy. At each level, blue squares represent the input nodes and colored dots refer to the peak nodes which are grouped from the intermediate clusters (connected-components). Note that the peaks at each level then become the input nodes at the next level. The right column shows the images corresponding to the peaks at each level of the three classes. The three classes converge at different levels: nodes of the red class already converges at the first level, the green class converge at level2 while the yellow class converge at level3. Best viewed in color.}
                \label{fig:hannah_deeper}
    \end{center}
\end{figure*}

\section{Experiment Details}

Here we describe additional experiment details including dataset statistics, input feature dimensions, sensitivity tests on \HNAME hyper-parameters, and the runtime hardware and software specifications.

\begin{table}[tbh]
\begin{center}
\begin{footnotesize}
\centering
    \begin{adjustbox}{max width=\linewidth}
            \begin{tabular}{c|c |c | c}
        \hline
    	Dataset & Images & Entities & Mean Cluster Size \\
    	\hline
    	\hline
    	TrillionPairs-Train \cite{TrillionPairs} & 669,560 & 18,084 & 37.0 \\ 
    	
    	Hannah-Test \cite{Hannah} & 201,240 & 251 & 801.8 \\
    	
    	IMDB-Test \cite{IMDB} & 1,265,173 & 50,289 & 25.2 \\
    	
    	IMDB-Test-SameDist \cite{IMDB} & 614,002 & 18,084 & 34.0 \\
    	
    	iNat2018-Train \cite{van2018inaturalist} & 324,418 & 5,690 & 57.0 \\
    	
    	iNat2018-Train-DifferentDist \cite{van2018inaturalist} &  51,696 & 5,690 & 9.0 \\
    	
    	iNat2018-Test \cite{van2018inaturalist} & 135,660 & 2,452 & 55.3 \\
    	
    	\hline
    \end{tabular}
    \end{adjustbox}
    \caption{\label{tbl:dataset_stats} Statistics of All Datasets }
\end{footnotesize}
\vspace{-1em}
\end{center}
\end{table}

\noindent\textbf{Dataset Statistics} \mytabref{tbl:dataset_stats} shows the detailed dataset statistics for all train and test sets used for the experiments.

\begin{table}[tbh]
\begin{center}
\begin{footnotesize}
\centering
     \begin{adjustbox}{max width=\linewidth}
            \begin{tabular}{c | c | c }
        \hline
    	 Hannah & IMDB & iNat2018   \\
    	 \hline
         128 & 128 & 512  \\
         \hline
    \end{tabular}
    \end{adjustbox}
    \caption{\label{tbl:input_dims} Input Feature Dimensions For All Datasets. }
\end{footnotesize}
\end{center}
\vspace{-1em}
\end{table}

\noindent\textbf{Input Feature} \mytabref{tbl:input_dims} lists the input feature dimensions for all datasets. $L_2$-normalization is applied on the features before network inference.
% \vspace{-0.5em}
% \noindent\textbf{Model Architecture and Losses} The \HNAME consists of the GNN module and the head to produce link approximation. We use GAT \cite{GAT} as the GNN module with hidden dimension 512 and 1-head attention for all the benchmarks. The link approximation head is a 2-layer MLP with 512 neurons on each layer. The weights for GAT and MLP are shared across all levels.

% When we compute the connectivity loss, we balance the positive and negative ratio of the edges by randomly down sample the larger part. For the final loss, we just sum up the balanced connectivity loss and density loss without further tuning the weight between them.
% % \vspace{-0.5em}

\noindent\textbf{Sensitivity Analysis over Hyper-parameters} ~\myfigref{fig:k_sensitivity} shows the sensitivity of \HNAME to the various hyper-parameters of the method including $k$ for $k$-NN build, $p_\tau$ for edge set decoding, the feature aggregation function choice detailed in Section 3.4 of the main paper, and the encoder layer architecture choice (GAT versus a vanilla GCN layer), mentioned in Section 3.3 of the main paper. The top two plots show the sensitivity of $k$, $p_\tau$ and the feature aggregation mechanism, where solid lines refer to identity feature aggregation and dashed lines represent the concatenation of identity and average feature. The bottom two plots show the sensitivity to the two different types of encoder layer architecture, a GAT (solid lines) and a vanilla GCN layer (dotted lines). Based on the validation set (a part of the meta-training set), the optimal hyper-parameters over the face clustering task are chosen as $k$ = 10, $p_\tau$ = 0.8, aggregation using identity feature only and encoding using GAT. Thus, for $k$ sensitivity, we vary it from 8 to 12. For $p_\tau$ sensitivity, we vary it from 0.75 to 0.85 with interval of 0.025. Metrics of NMI (blue), Fp (yellow), and Fb (red) are shown. The plots show that varying $k$ and $p_\tau$ near the optimal value does not result in significant changes in results. The differences in final clustering accuracy between identity-feature-only aggregation and concatenation of both identity and average feature, as well as the variations between using GAT versus a vanilla GCN layer in encoding, are small.

%  \yifan{Do we need to show ablation on the data augmentation? to prevent reviewer mentioning the generalization comes just from the augmentation itself. Put the detailed parameters into supplementary.}\david{It could be worth putting the detailed parameters into the supplementary, especially because space will likely be limited.  The only exception is the $k=10$ setting for all experiments.  This might be nice to mention for marketing purposes.}

% This augmentation creates more hierarchies which leads to more training time, as the smaller $k$ requires more levels to cover the whole cluster, though the number of levels $L$ converges exponentially. Assume that the cluster is a $k$-ary tree, then the selection of $L$ follows the formula: $L = \log_k {(C \cdot k-C+1)} - 1$ where $C$ is the average cluster size for a dataset. Based on the formula and the $k$ values, we use $L=2, 3, 4$ for face and nature benchmarks and $L=1, 2, 3$ for fashion benchmarks. Note that $k=5, 3$ are only used for training augmentation, for test-time inference we always use $k=10$ and iterate until the early-stopping criteria is met. 

% For training, we apply data augmentation by building multiple sets of hierarchies of $k$-NN graphs with $k=10, 5, 3$ and levels of $l=2, 3, 4$ together.

\noindent\textbf{Additional Training Details} For the base clustering model \NAME, we use 1 layer of GAT as encoder and a 2-layer MLP for joint linkage and density prediction. Both face and nature species models are trained for 250 epochs with batchsize 4096. All models use SGD optimizer with 0.1 base learning rate, 0.9 momentum, and 1e-5 weight decay. The learning rate follows a cosine annealing schedule \cite{cosine}.

\noindent\textbf{Runtime Experiment Hardware and Software} We measure the runtime (Section 4.7 of the main paper) with 8-core Intel(R) Xeon(R) E5-2686 v4 CPU and Tesla V100 GPU. Our models use PyTorch\cite{pytorch} v1.5, DGL\cite{dgl} v0.6 with CUDA v10.1. $k$-NN building leverages faiss\cite{faiss}.

{\small
\bibliographystyle{ieee_fullname}
\bibliography{graph-clustering.bib}

\begin{thebibliography}{10}\itemsep=-1pt

\bibitem{nyupaper}
\url{https://cs.nyu.edu/media/publications/choma_nicholas.pdf}.

\bibitem{TrillionPairs}
\url{http://trillionpairs.deepglint.com/overview}.

\bibitem{OPTICS}
Mihael Ankerst, Markus~M Breunig, Hans-Peter Kriegel, and J{\"o}rg Sander.
\newblock Optics: ordering points to identify the clustering structure.
\newblock {\em ACM Sigmod record}, 28(2):49--60, 1999.

\bibitem{bianchi2019hierarchical}
Filippo~Maria Bianchi, Daniele Grattarola, Lorenzo Livi, and Cesare Alippi.
\newblock Hierarchical representation learning in graph neural networks with
  node decimation pooling.
\newblock {\em arXiv preprint arXiv:1910.11436}, 2019.

\bibitem{bonald2018hierarchical}
Thomas Bonald, Bertrand Charpentier, Alexis Galland, and Alexandre Hollocou.
\newblock Hierarchical graph clustering using node pair sampling.
\newblock {\em arXiv preprint arXiv:1806.01664}, 2018.

\bibitem{braso2020learning}
Guillem Bras{\'o} and Laura Leal-Taix{\'e}.
\newblock Learning a neural solver for multiple object tracking.
\newblock In {\em Proceedings of the IEEE/CVF Conference on Computer Vision and
  Pattern Recognition}, pages 6247--6257, 2020.

\bibitem{brown2020smooth}
Andrew Brown, Weidi Xie, Vicky Kalogeiton, and Andrew Zisserman.
\newblock Smooth-ap: Smoothing the path towards large-scale image retrieval.
\newblock {\em arXiv preprint arXiv:2007.12163}, 2020.

\bibitem{H-DBSCAN}
Ricardo~JGB Campello, Davoud Moulavi, and J{\"o}rg Sander.
\newblock Density-based clustering based on hierarchical density estimates.
\newblock In {\em Pacific-Asia conference on knowledge discovery and data
  mining}, pages 160--172. Springer, 2013.

\bibitem{chen2017harp}
Haochen Chen, Bryan Perozzi, Yifan Hu, and Steven Skiena.
\newblock Harp: Hierarchical representation learning for networks.
\newblock {\em arXiv preprint arXiv:1706.07845}, 2017.

\bibitem{FASTGCN}
Jie Chen, Tengfei Ma, and Cao Xiao.
\newblock Fastgcn: fast learning with graph convolutional networks via
  importance sampling.
\newblock {\em arXiv preprint arXiv:1801.10247}, 2018.

\bibitem{chen2019graph}
Yunpeng Chen, Marcus Rohrbach, Zhicheng Yan, Yan Shuicheng, Jiashi Feng, and
  Yannis Kalantidis.
\newblock Graph-based global reasoning networks.
\newblock In {\em Proceedings of the IEEE Conference on Computer Vision and
  Pattern Recognition}, pages 433--442, 2019.

\bibitem{defferrard2016convolutional}
Micha{\"e}l Defferrard, Xavier Bresson, and Pierre Vandergheynst.
\newblock Convolutional neural networks on graphs with fast localized spectral
  filtering.
\newblock In {\em Advances in neural information processing systems}, pages
  3844--3852, 2016.

\bibitem{Graclus}
Inderjit~S Dhillon, Yuqiang Guan, and Brian Kulis.
\newblock Weighted graph cuts without eigenvectors a multilevel approach.
\newblock {\em IEEE transactions on pattern analysis and machine intelligence},
  29(11):1944--1957, 2007.

\bibitem{DB-SCAN}
Martin Ester, Hans-Peter Kriegel, J{\"o}rg Sander, Xiaowei Xu, et~al.
\newblock A density-based algorithm for discovering clusters in large spatial
  databases with noise.

\bibitem{h_2D_3D_match}
Mohammed~E Fathy, Quoc-Huy Tran, M Zeeshan~Zia, Paul Vernaza, and Manmohan
  Chandraker.
\newblock Hierarchical metric learning and matching for 2d and 3d geometric
  correspondences.
\newblock In {\em Proceedings of the European Conference on Computer Vision
  (ECCV)}, pages 803--819, 2018.

\bibitem{Graph-SAGE}
Will Hamilton, Zhitao Ying, and Jure Leskovec.
\newblock Inductive representation learning on large graphs.
\newblock In {\em Advances in neural information processing systems}, pages
  1024--1034, 2017.

\bibitem{Spectral}
J. {Ho}, {Ming-Husang Yang}, {Jongwoo Lim}, {Kuang-Chih Lee}, and D.
  {Kriegman}.
\newblock Clustering appearances of objects under varying illumination
  conditions.
\newblock In {\em 2003 IEEE Computer Society Conference on Computer Vision and
  Pattern Recognition, 2003. Proceedings.}, volume~1, pages I--I, 2003.

\bibitem{hu2019hierarchical}
Fenyu Hu, Yanqiao Zhu, Shu Wu, Liang Wang, and Tieniu Tan.
\newblock Hierarchical graph convolutional networks for semi-supervised node
  classification.
\newblock {\em arXiv preprint arXiv:1902.06667}, 2019.

\bibitem{huang2019attpool}
Jingjia Huang, Zhangheng Li, Nannan Li, Shan Liu, and Ge Li.
\newblock Attpool: Towards hierarchical feature representation in graph
  convolutional networks via attention mechanism.
\newblock In {\em Proceedings of the IEEE International Conference on Computer
  Vision}, pages 6480--6489, 2019.

\bibitem{faiss}
Jeff Johnson, Matthijs Douze, and Herv{\'e} J{\'e}gou.
\newblock Billion-scale similarity search with gpus.
\newblock {\em IEEE Transactions on Big Data}, 2019.

\bibitem{kipf2016semi}
Thomas~N Kipf and Max Welling.
\newblock Semi-supervised classification with graph convolutional networks.
\newblock {\em arXiv preprint arXiv:1609.02907}, 2016.

\bibitem{h_future_action}
Tian Lan, Tsung-Chuan Chen, and Silvio Savarese.
\newblock A hierarchical representation for future action prediction.
\newblock In {\em European Conference on Computer Vision}, pages 689--704.
  Springer, 2014.

\bibitem{fpn_object}
Tsung-Yi Lin, Piotr Dollar, Ross Girshick, Kaiming He, Bharath Hariharan, and
  Serge Belongie.
\newblock Feature pyramid networks for object detection.
\newblock In {\em Proceedings of the IEEE Conference on Computer Vision and
  Pattern Recognition (CVPR)}, July 2017.

\bibitem{lin2018deep}
Wei-An Lin, Jun-Cheng Chen, Carlos~D Castillo, and Rama Chellappa.
\newblock Deep density clustering of unconstrained faces.
\newblock In {\em Proceedings of the IEEE Conference on Computer Vision and
  Pattern Recognition}, pages 8128--8137, 2018.

\bibitem{lin2017proximity}
Wei-An Lin, Jun-Cheng Chen, and Rama Chellappa.
\newblock A proximity-aware hierarchical clustering of faces.
\newblock In {\em 2017 12th IEEE International Conference on Automatic Face \&
  Gesture Recognition (FG 2017)}, pages 294--301. IEEE, 2017.

\bibitem{lipov2020multiscale}
Alex Lipov and Pietro Li{\`o}.
\newblock A multiscale graph convolutional network using hierarchical
  clustering.
\newblock {\em arXiv preprint arXiv:2006.12542}, 2020.

\bibitem{liu2017sphereface}
Weiyang Liu, Yandong Wen, Zhiding Yu, Ming Li, Bhiksha Raj, and Le Song.
\newblock Sphereface: Deep hypersphere embedding for face recognition.
\newblock In {\em Proceedings of the IEEE conference on computer vision and
  pattern recognition}, pages 212--220, 2017.

\bibitem{k-means}
Stuart Lloyd.
\newblock Least squares quantization in pcm.
\newblock {\em IEEE transactions on information theory}, 28(2):129--137, 1982.

\bibitem{h_mid_top}
Hans Lobel, Ren{\'e} Vidal, and Alvaro Soto.
\newblock Hierarchical joint max-margin learning of mid and top level
  representations for visual recognition.
\newblock In {\em Proceedings of the IEEE International Conference on Computer
  Vision}, pages 1697--1704, 2013.

\bibitem{cosine}
Ilya Loshchilov and Frank Hutter.
\newblock Sgdr: Stochastic gradient descent with warm restarts.
\newblock {\em arXiv preprint arXiv:1608.03983}, 2016.

\bibitem{h_track}
Chao Ma, Jia-Bin Huang, Xiaokang Yang, and Ming-Hsuan Yang.
\newblock Hierarchical convolutional features for visual tracking.
\newblock In {\em Proceedings of the IEEE international conference on computer
  vision}, pages 3074--3082, 2015.

\bibitem{IJBC}
Brianna Maze, Jocelyn Adams, James~A Duncan, Nathan Kalka, Tim Miller, Charles
  Otto, Anil~K Jain, W~Tyler Niggel, Janet Anderson, Jordan Cheney, et~al.
\newblock Iarpa janus benchmark-c: Face dataset and protocol.
\newblock In {\em 2018 International Conference on Biometrics (ICB)}, pages
  158--165. IEEE, 2018.

\bibitem{hierarchical-VRD}
Li Mi and Zhenzhong Chen.
\newblock Hierarchical graph attention network for visual relationship
  detection.
\newblock In {\em Proceedings of the IEEE/CVF Conference on Computer Vision and
  Pattern Recognition}, pages 13886--13895, 2020.

\bibitem{newman2004fast}
Mark~EJ Newman.
\newblock Fast algorithm for detecting community structure in networks.
\newblock {\em Physical review E}, 69(6):066133, 2004.

\bibitem{ng2002spectral}
Andrew~Y Ng, Michael~I Jordan, and Yair Weiss.
\newblock On spectral clustering: Analysis and an algorithm.
\newblock In {\em Advances in neural information processing systems}, pages
  849--856, 2002.

\bibitem{h_vision_language_task}
Duy-Kien Nguyen and Takayuki Okatani.
\newblock Multi-task learning of hierarchical vision-language representation.
\newblock In {\em Proceedings of the IEEE Conference on Computer Vision and
  Pattern Recognition}, pages 10492--10501, 2019.

\bibitem{ARO}
Charles Otto, Dayong Wang, and Anil~K Jain.
\newblock Clustering millions of faces by identity.
\newblock {\em IEEE transactions on pattern analysis and machine intelligence},
  40(2):289--303, 2017.

\bibitem{Hannah}
Alexey Ozerov, Jean-Ronan Vigouroux, Louis Chevallier, and Patrick P{\'e}rez.
\newblock On evaluating face tracks in movies.
\newblock In {\em 2013 IEEE International Conference on Image Processing},
  pages 3003--3007. IEEE, 2013.

\bibitem{pytorch}
Adam Paszke, Sam Gross, Francisco Massa, Adam Lerer, James Bradbury, Gregory
  Chanan, Trevor Killeen, Zeming Lin, Natalia Gimelshein, Luca Antiga, et~al.
\newblock Pytorch: An imperative style, high-performance deep learning library.
\newblock In {\em Advances in neural information processing systems}, pages
  8026--8037, 2019.

\bibitem{pons2005computing}
Pascal Pons and Matthieu Latapy.
\newblock Computing communities in large networks using random walks.
\newblock In {\em International symposium on computer and information
  sciences}, pages 284--293. Springer, 2005.

\bibitem{roychowdhury2020improving}
Aruni RoyChowdhury, Xiang Yu, Kihyuk Sohn, Erik Learned-Miller, and Manmohan
  Chandraker.
\newblock Improving face recognition by clustering unlabeled faces in the wild.
\newblock In {\em European Conference on Computer Vision}, pages 119--136.
  Springer, 2020.

\bibitem{FINCH}
Saquib Sarfraz, Vivek Sharma, and Rainer Stiefelhagen.
\newblock Efficient parameter-free clustering using first neighbor relations.
\newblock In {\em Proceedings of the IEEE/CVF Conference on Computer Vision and
  Pattern Recognition}, pages 8934--8943, 2019.

\bibitem{towardopenset}
Walter~J Scheirer, Anderson de Rezende~Rocha, Archana Sapkota, and Terrance~E
  Boult.
\newblock Toward open set recognition.
\newblock {\em IEEE transactions on pattern analysis and machine intelligence},
  35(7):1757--1772, 2012.

\bibitem{HAC}
Robin Sibson.
\newblock Slink: an optimally efficient algorithm for the single-link cluster
  method.
\newblock {\em The computer journal}, 16(1):30--34, 1973.

\bibitem{slonim1999agglomerative}
Noam Slonim and Naftali Tishby.
\newblock Agglomerative information bottleneck.
\newblock In {\em NIPS}, volume~4, 1999.

\bibitem{vanDerMaaten2008}
Laurens van~der Maaten and Geoffrey Hinton.
\newblock Visualizing data using {t-SNE}.
\newblock {\em Journal of Machine Learning Research}, 9:2579--2605, 2008.

\bibitem{van2018inaturalist}
Grant Van~Horn, Oisin Mac~Aodha, Yang Song, Yin Cui, Chen Sun, Alex Shepard,
  Hartwig Adam, Pietro Perona, and Serge Belongie.
\newblock The inaturalist species classification and detection dataset.
\newblock In {\em Proceedings of the IEEE conference on computer vision and
  pattern recognition}, pages 8769--8778, 2018.

\bibitem{quickshift}
Andrea Vedaldi and Stefano Soatto.
\newblock Quick shift and kernel methods for mode seeking.
\newblock In {\em European conference on computer vision}, pages 705--718.
  Springer, 2008.

\bibitem{GAT}
Petar Veli{\v{c}}kovi{\'c}, Guillem Cucurull, Arantxa Casanova, Adriana Romero,
  Pietro Lio, and Yoshua Bengio.
\newblock Graph attention networks.
\newblock {\em arXiv preprint arXiv:1710.10903}, 2017.

\bibitem{NMI}
Nguyen~Xuan Vinh, Julien Epps, and James Bailey.
\newblock Information theoretic measures for clusterings comparison: Variants,
  properties, normalization and correction for chance.
\newblock {\em The Journal of Machine Learning Research}, 11:2837--2854, 2010.

\bibitem{von2007tutorial}
Ulrike Von~Luxburg.
\newblock A tutorial on spectral clustering.
\newblock {\em Statistics and computing}, 17(4):395--416, 2007.

\bibitem{IMDB}
Fei Wang, Liren Chen, Cheng Li, Shiyao Huang, Yanjie Chen, Chen Qian, and
  Chen~Change Loy.
\newblock The devil of face recognition is in the noise.
\newblock {\em arXiv preprint arXiv:1807.11649}, 2018.

\bibitem{arcface}
Feng Wang, Jian Cheng, Weiyang Liu, and Haijun Liu.
\newblock Additive margin softmax for face verification.
\newblock {\em IEEE Signal Processing Letters}, 25(7):926--930, 2018.

\bibitem{dgl}
Minjie Wang, Lingfan Yu, Da Zheng, Quan Gan, Yu Gai, Zihao Ye, Mufei Li,
  Jinjing Zhou, Qi Huang, Chao Ma, et~al.
\newblock Deep graph library: Towards efficient and scalable deep learning on
  graphs.
\newblock {\em arXiv preprint arXiv:1909.01315}, 2019.

\bibitem{EdgeConv}
Yue Wang, Yongbin Sun, Ziwei Liu, Sanjay~E Sarma, Michael~M Bronstein, and
  Justin~M Solomon.
\newblock Dynamic graph cnn for learning on point clouds.
\newblock {\em Acm Transactions On Graphics (tog)}, 38(5):1--12, 2019.

\bibitem{L-GCN}
Z. {Wang}, L. {Zheng}, Y. {Li}, and S. {Wang}.
\newblock Linkage based face clustering via graph convolution network.
\newblock In {\em 2019 IEEE/CVF Conference on Computer Vision and Pattern
  Recognition (CVPR)}, pages 1117--1125, 2019.

\bibitem{weng2020gnn3dmot}
Xinshuo Weng, Yongxin Wang, Yunze Man, and Kris Kitani.
\newblock Gnn3dmot: Graph neural network for 3d multi-object tracking with
  multi-feature learning.
\newblock {\em arXiv preprint arXiv:2006.07327}, 2020.

\bibitem{h_finegrain}
Lingxi Xie, Qi Tian, Richang Hong, Shuicheng Yan, and Bo Zhang.
\newblock Hierarchical part matching for fine-grained visual categorization.
\newblock In {\em Proceedings of the IEEE International Conference on Computer
  Vision}, pages 1641--1648, 2013.

\bibitem{yan2019convolutional}
Sijie Yan, Zhizhong Li, Yuanjun Xiong, Huahan Yan, and Dahua Lin.
\newblock Convolutional sequence generation for skeleton-based action
  synthesis.
\newblock In {\em Proceedings of the IEEE International Conference on Computer
  Vision}, pages 4394--4402, 2019.

\bibitem{yan2018spatial}
Sijie Yan, Yuanjun Xiong, and Dahua Lin.
\newblock Spatial temporal graph convolutional networks for skeleton-based
  action recognition.
\newblock {\em arXiv preprint arXiv:1801.07455}, 2018.

\bibitem{GCN-VE}
Lei Yang, Dapeng Chen, Xiaohang Zhan, Rui Zhao, Chen~Change Loy, and Dahua Lin.
\newblock Learning to cluster faces via confidence and connectivity estimation.
\newblock In {\em Proceedings of the IEEE/CVF Conference on Computer Vision and
  Pattern Recognition}, pages 13369--13378, 2020.

\bibitem{GCN-DS}
Lei Yang, Xiaohang Zhan, Dapeng Chen, Junjie Yan, Chen~Change Loy, and Dahua
  Lin.
\newblock Learning to cluster faces on an affinity graph.
\newblock In {\em Proceedings of the IEEE Conference on Computer Vision and
  Pattern Recognition}, pages 2298--2306, 2019.

\bibitem{ying2018hierarchical}
Zhitao Ying, Jiaxuan You, Christopher Morris, Xiang Ren, Will Hamilton, and
  Jure Leskovec.
\newblock Hierarchical graph representation learning with differentiable
  pooling.
\newblock In {\em Advances in neural information processing systems}, pages
  4800--4810, 2018.

\bibitem{CDP}
Xiaohang Zhan, Ziwei Liu, Junjie Yan, Dahua Lin, and Chen~Change Loy.
\newblock Consensus-driven propagation in massive unlabeled data for face
  recognition.
\newblock In {\em Proceedings of the European Conference on Computer Vision
  (ECCV)}, September 2018.

\bibitem{zhu2011rank}
Chunhui Zhu, Fang Wen, and Jian Sun.
\newblock A rank-order distance based clustering algorithm for face tagging.
\newblock In {\em CVPR 2011}, pages 481--488. IEEE, 2011.

\bibitem{persistent_homology}
Afra Zomorodian and Gunnar Carlsson.
\newblock Computing persistent homology.
\newblock {\em Discrete \& Computational Geometry}, 33(2):249--274, 2005.

\end{thebibliography}
}

\end{document}